\documentclass{article}

\usepackage{microtype}
\usepackage{graphicx}
\usepackage{subcaption}
\usepackage{booktabs} 

\usepackage[dvipsnames]{xcolor}

\usepackage{booktabs}
\usepackage{multirow}
\usepackage{graphicx} 
\usepackage{makecell}
\usepackage{siunitx}
\usepackage[table]{xcolor}

\usepackage[preprint]{icml2026}

\usepackage{amsmath}
\usepackage{amssymb}
\usepackage{mathtools}
\usepackage{amsthm}
\usepackage{enumitem}

\usepackage{tcolorbox}

\usepackage[capitalize,noabbrev]{cleveref}

\theoremstyle{plain}

\theoremstyle{definition}

\theoremstyle{remark}

\usepackage[textsize=tiny]{todonotes}

\icmltitlerunning{Learning to Detect Language Model Training Data via Active Reconstruction}

\begin{document}

\twocolumn[
  \icmltitle{
Learning to Detect Language Model Training Data via Active Reconstruction

}



  \icmlsetsymbol{equal}{*}

  \begin{icmlauthorlist}
    \icmlauthor{Junjie Oscar Yin}{uw}
    \icmlauthor{John X. Morris}{cornell}
    \icmlauthor{Vitaly Shmatikov}{cornell}
    \icmlauthor{Sewon Min}{ucb,ai2}
    \icmlauthor{Hannaneh Hajishirzi}{uw,ai2}
  \end{icmlauthorlist}

  \icmlaffiliation{uw}{University of Washington}
  \icmlaffiliation{ucb}{UC Berkeley}
  \icmlaffiliation{cornell}{Cornell University}
  \icmlaffiliation{ai2}{Allen Institute for Artificial Intelligence}

  \icmlcorrespondingauthor{Junjie Oscar Yin}{osey@cs.washington.edu}

  \icmlkeywords{Machine Learning, ICML}

  \vskip 0.3in
]

\newcommand{\fix}{\marginpar{FIX}}
\newcommand{\new}{\marginpar{NEW}}

\newcommand{\Jack}[1]{\textcolor{red}{[Jack: #1]}}
\newcommand{\Oscar}[1]{\textcolor{blue}{[Oscar: #1]}}
\newcommand{\Sewon}[1]{\textcolor{orange}{[Sewon: #1]}}
\newcommand{\Vitaly}[1]{\textcolor{teal}{[Vitaly: #1]}}
\newcommand{\hanna}[1]{\textcolor{purple}{[Hanna: #1]}}

\newcommand{\Method}{\textsc{ADRA}}
\newcommand{\MethodPlus}{\textsc{ADRA+}}

\definecolor{hdrblue}{RGB}{224,234,249}   
\definecolor{hdrpink}{RGB}{255,232,240}   
\newcommand{\na}{\multicolumn{1}{c}{--}}
\newcommand{\sect}[1]{\rowcolor{gray!18}\multicolumn{9}{l}{\textbf{#1}}\\}
\newcommand{\rowgap}{\addlinespace[0.45ex]} 
\newcommand{\bestavg}[2]{{\scriptsize #1} / {#2}}

\newcommand{\sectt}[1]{\rowcolor{gray!18}\multicolumn{10}{l}{\textbf{#1}}\\}



\printAffiliationsAndNotice{}  

\begin{abstract}
Detecting LLM training data is generally framed as a membership inference attack (MIA) problem. However, conventional MIAs operate passively on fixed model weights, using log-likelihoods or text generations. In this work, we introduce \textbf{Active Data Reconstruction Attack} (ADRA), a family of MIA that actively induces a model to reconstruct a given text through training. We hypothesize that training data are \textit{more reconstructible} than non-members, and the difference in their reconstructibility can be exploited for membership inference. Motivated by findings that reinforcement learning (RL) sharpens behaviors already encoded in weights, we leverage on-policy RL to actively elicit data reconstruction by finetuning a policy initialized from the target model. To effectively use RL for MIA, we design reconstruction metrics and contrastive rewards. The resulting algorithms, \textsc{ADRA} and its adaptive variant \textsc{ADRA+}, improve both reconstruction and detection given a pool of candidate data.
Experiments show that our methods consistently outperform existing MIAs in detecting pre-training, post-training, and distillation data, with an average improvement of 10.7\% over the previous runner-up.  In particular, \MethodPlus~improves over Min-K\%++ by 18.8\% on BookMIA for pre-training detection and by 7.6\% on AIME for post-training detection.

\end{abstract}

\vspace{-0.16in}
\section{Introduction}

Modern language models are known to memorize and regurgitate training data, raising concerns about copyright, privacy, and data contamination \citep{carlini2021extracting, ahmed2026extracting}. Membership inference attacks (MIAs) are methods that detect whether a given input was present in the training data \citep{shokri2017membership}, and have shown some success when applied to LLMs \citep{shi2023detecting}.

%

\begin{figure}[!t]
    \centering
    \resizebox{\linewidth}{!}{%
    \includegraphics{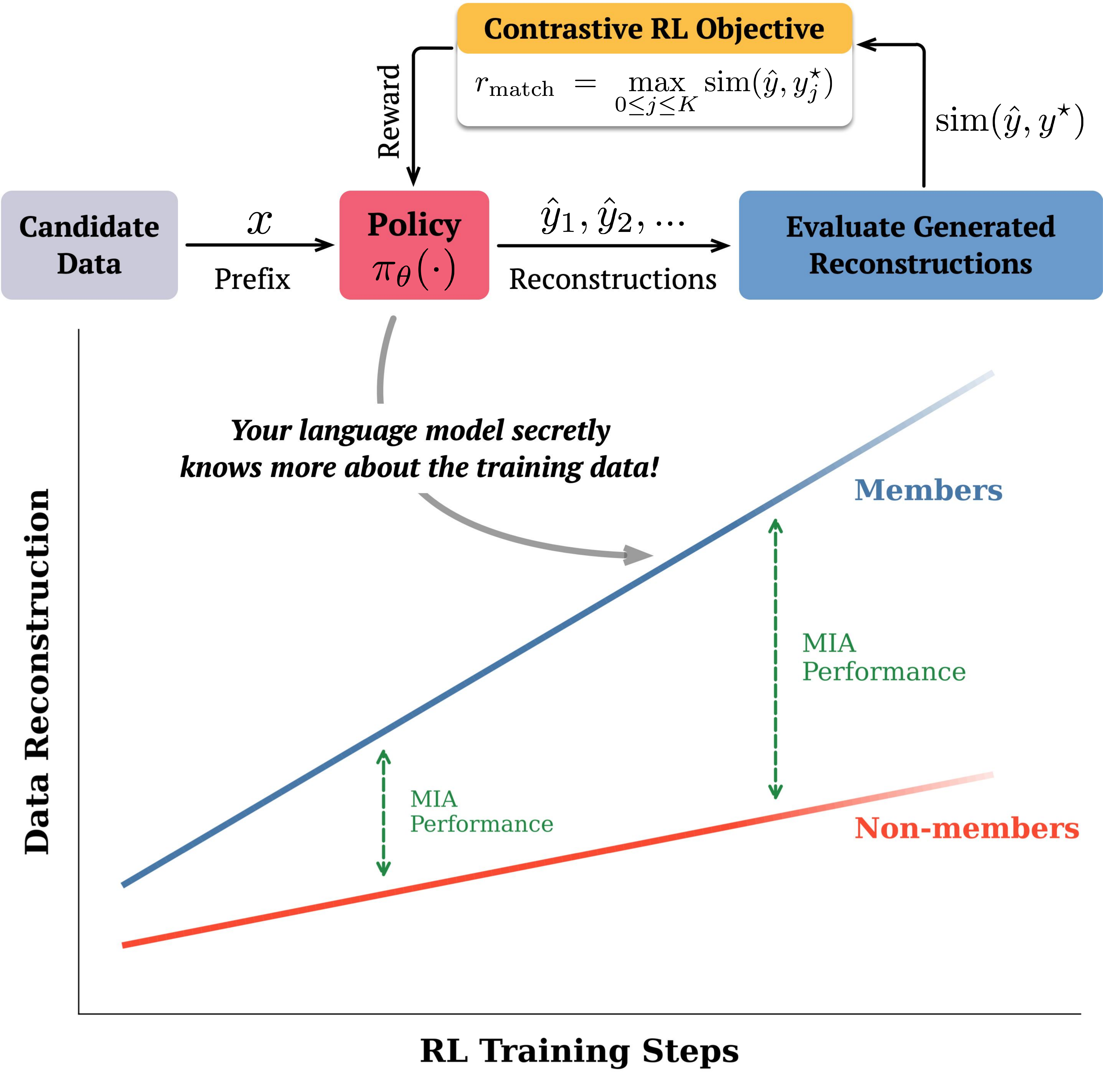} 
    }
    \caption{\textbf{Active Data Reconstruction Attack.} Language model generates reconstructions from a candidate prefix and is rewarded via a contrastive objective. Members become easier to reconstruct than non-members over RL training, improving MIA performance.}
    \label{fig:overview}
    \vspace{-0.2in}
\end{figure}
%


Prior MIAs query the target model without modifying its weights, \textit{passively} extracting signals exposed through the output layer---such as loss \citep{shokri2017membership,yeom2018privacy, carlini2021extracting}, log-probabilities \citep{shi2023detecting,zhang2024min}, or text generations \citep{hallinan2025surprising}. We hypothesize that model weights  encode \textit{latent membership signal} that passive methods fail to reliably expose. If such a signal exists, it could be \textit{actively} elicited through model weight updates.

We introduce \textbf{Active Data Reconstruction Attack (ADRA)}, a family of MIAs that uses Reinforcement Learning (RL) with reconstruction rewards to elicit latent membership signals. Because model weights encode traces of training data, RL training exploits the difference in member and non-member reconstructability for detection. Our approach is motivated by recent findings that RL sharpens behaviors already encoded in weights \citep{yue2025does, wang2025reinforcement} while transferring minimal new information \citep{schulman2025lora}, suggesting RL is well-suited to surface latent reconstruction signal. 


\Cref{fig:overview} illustrates our attack: given a candidate prefix, the model generates continuations (\textit{reconstructions}) and is rewarded for producing generations similar to the candidate suffix. 
To effectively leverage RL for MIA, we design reconstruction metrics and contrastive rewards that match generations against a pool containing the true suffix and negative distractors. This yields two algorithms: ADRA, which rewards the best match among the pool, and ADRA+, which adaptively adjusts how often the true suffix should be included during matching based on a prior derived from loss-based scores. 





Empirically, we run comprehensive experiments spanning pre-training, post-training, and distillation settings across 6 open-weight LLMs. To better match frontier models' knowledge cutoffs and heavier post-training, we construct 6 new MIA datasets and evaluate on 2 established benchmarks. We find that \Method~and \MethodPlus~\emph{consistently outperform prior MIAs across all settings}, with an average improvement of 10.7\% over the previous runner-up.
For pre-training data detection, on the challenging WikiMIA\textsubscript{2024} Hard benchmark—where most baselines hover below random—\MethodPlus~reaches 60.6\% AUROC, outperforming Min-K\%++ by 10\%. For post-training, under controlled contamination on AIME, \MethodPlus~achieves 85.9\% AUROC, improving over N-Sampling by 13.2\% and Min-K\%++ by 7.6\%. For distillation, on prefix-matched distillation traces, \Method~attains near-perfect membership inference, reaching 98.4\% under Deepseek-R1 distillation. Ablations confirm that RL-based optimization, reconstruction-reward design, and the contrastive objective are critical to these gains.

Our results suggest that model weights encode more about training data than fixed outputs reveal. RL training help surface this latent information, enabling stronger reconstruction and membership inference. Codebase, models, and datasets are available at https://github.com/oseyosey/MIA-RL.

\section{Background}
\label{sec:background}

Membership inference \citep{shokri2017membership} aims to determine if a datapoint $x$ was present in the training data of model $\theta$. We focus on the specific problem of detecting training data from LLMs, which is known to be difficult \citep{duan2024membership}.
\vspace{-0.15in}
\label{sec:formalization}

\paragraph{Notation.}
Let $p_\theta$ be the target LLM with parameters $\theta$, trained on a dataset $\mathcal{D}_{train}$.  
Given a candidate document $x\in\mathcal{D}_{cand}$, write $x=(c,\,y^\star)$ where $c$ is the observed prefix and $y^\star$ is the held-out suffix. 
Membership inference is the binary classification problem of determining whether $x$ was drawn from $\mathcal{D}_{train}$ or not. Define a scoring function $f(x; \theta)$ that maps each candidate to a scalar membership score; MIAs predict ``member'' iff $f(x;\theta)\ge \epsilon$ for a threshold $\epsilon$.

\vspace{.1cm}\noindent\textbf{Loss-based Methods.} Most prior MIAs score a candidate by the model's token-level log-probabilities on $x$, with the intuition that models are likely to have a lower loss on sequences that have been seen during training:
\vspace{-0.10in}
\begin{itemize}[itemsep=1mm,leftmargin=4mm]
    \item \textit{Loss}~\citep{yeom2018privacy}: $f(x)=\mathcal{L}(x;\theta)$.
    \item \textit{R-Loss}~\citep{carlini2021extracting}: calibrate by a reference model $\theta_{\mathrm{ref}}$, \; $f=\mathcal{L}(x;\theta)-\mathcal{L}(x;\theta_{\mathrm{ref}})$.
    \item \textit{Zlib Entropy}~\citep{carlini2021extracting}: normalize by the zlib compression size of $x$, \; $f=\mathcal{L}(x;\theta)/\mathrm{zlib}(x)$.
    \item \textit{Min-$K\%$}~\citep{shi2023detecting}: average over the $k\%$ lowest-likelihood tokens, \; $f=\frac{1}{|k(x)|}\sum_{i\in\mathrm{min}\;k(x)} -\log p(x_i\mid x_{<i};\theta)$.
    \item \textit{Min-$K\%$++}~\citep{zhang2024min}: z-score each next-token log-prob by the vocab mean/std under $\theta$, then average the $k\%$ smallest: $s_i=\frac{\log p(x_i\mid x_{<i};\theta)-\mu_{x_{<i}}}{\sigma_{x_{<i}}}$, $\; f=\frac{1}{|k(x)|}\sum_{i\in\mathrm{min}\;k(x)} s_i$.
\end{itemize}

\vspace{.1cm}\noindent\textbf{Reconstruction-based Methods.} 
 \citet{hallinan2025surprising} propose sampling completions and scoring membership via n-gram metrics---effectively a reconstruction-based MIA. Formally, given a fixed prompt $q$ and prefix $c$, we sample $N$ completions $\{\hat{y}\}^{N}_{i=1}$ from $p_\theta$ with nucleus sampling:
\begin{equation}
\{\hat{y}^{(i)}\}^{N}_{i=1}\sim p_\theta(\cdot\mid q, c).
\end{equation}
We score each candidate $x$ by comparing each completion $\hat{y}^{(i)}$ to the ground truth suffix $y^\star$ using a similarity function $\textrm{sim}(\hat{y}^{(i)}, y^\star)$:
\begin{equation}
\label{eq:score}
f(x)\;=\;\frac{1}{N}\sum_{i=1}^{N} \textrm{sim}\big(\hat{y}^{(i)},y^\star\big).
\end{equation}
\section{Active Data Reconstruction Attack}
\label{sec:method}

The above-mentioned MIAs operate on fixed model weights, querying the target model's log-probabilities or generations. They assume that passive querying extracts all available membership signal. Conversely, we ask if training data leaves a \textit{latent membership signal} in a model's parameters that is not captured by existing MIA methods.

If such latent signal exists, we can design an \textit{Active MIA} that actively elicit the latent signal rather than passive measurements. In this work, we propose \textsc{Active Data Reconstruction Attack (ADRA)}.


We first describe our RL formulation in \Cref{sec:rl-induced-reconstruction}, then detail our reward design. Specifically, \Cref{sec:reconstructoin-rewards} defines similarity functions that reward reconstruction quality, and \Cref{sec:contrastive-rl-objective} specifies how we shape these rewards via contrastive structure to constrain supervision.


\subsection{RL-induced Data Reconstruction}
\label{sec:rl-induced-reconstruction}
Recent work shows that RL sharpens behaviors already encoded in weights \citep{yue2025does, wang2025reinforcement} while transferring minimal new information \citep{schulman2025lora}, suggesting RL could be well-suited to elicit latent membership signals. We leverage this insight by explicitly rewarding the model for reconstructing candidate data.

Concretely, we fine-tune a policy initialized from target model $p_{\theta}$ with on-policy RL, reusing the similarity function from \Cref{eq:score} as the reward:
\begin{equation}
\label{eq:Naive-Reward}
r^{(i)} \leftarrow \text{sim}(\hat{y}^{(i)}, y^\star), \quad \forall i \in \{1,...,N\}.
\end{equation}
For candidate $x$, the policy generates completions $\hat{y}$ given prefix $(q,c)$ and receives a scalar reward $r(\hat{y},x)=\textrm{sim}(\hat{y},\,y^\star)$. We update the policy by maximizing the expected return:
\begin{equation}
\label{eq:RL-Objective}
J_x(\theta)\;=\;\mathbb{E}_{\hat{y}\sim p_\theta(\cdot\mid q,c)}\!\big[r(\hat{y},x)\big].
\end{equation}

We estimate \Cref{eq:RL-Objective} with a Monte Carlo average over the $N$ samples above. In practice we apply Group Relative Policy Optimization (GRPO) \citep{shao2024deepseekmath}, which optimizes a clipped/KL-regularized surrogate of \Cref{eq:RL-Objective}. After $K$ on-policy RL updates, we obtain $\theta_K$. To evaluate $x$, we resample $M$ completions $\hat{y}^{(1:M)}\sim p_{\theta_K}(\cdot\mid c)$ and re-aggregate their similarities into a membership score.



\begin{algorithm}[t]
\caption{ \textsc{Active Data Reconstruction Attack} (single candidate $x=(c,y^\star)$; batched in practice).}
\label{alg:ddrl}
\begin{algorithmic}[1]

\INPUT target LM $p_\theta$, (optional) prompt $q$, candidate $x=(c,y^\star)$, threshold $\epsilon$, RL steps $K$, samples per step $N$, reward mode $\textsc{Mode}\in\{\textsc{Match},\textsc{Adapt}\}$, similarity $\mathrm{sim}(\cdot,\cdot)$.

\STATE Initialize $\theta_0 \leftarrow \theta$
\FOR{$k=0$ to $K-1$}
  \STATE Sample rollouts $\hat{y}^{(1{:}N)} \sim p_{\theta_k}(\cdot \mid q,c)$
  \STATE Compute rewards $r^{(i)} \leftarrow r_{\textsc{Mode}}(\hat{y}^{(i)},x)$ \COMMENT{$r_{\textsc{Match}}$ / $r_{\textsc{Adapt}}$}
  \STATE Update $\theta_{k+1}$ using GRPO/PPO (KL/clipping)
\ENDFOR
\STATE Evaluate $S(x)$ by computing $\mathrm{sim}\!\big(\hat{y}^{(i)},x\big)$ over new samples from $p_{\theta_K}(\cdot\mid c)$
\OUTPUT membership score $S(x)$, reconstructions $\hat{y}^{1:N}_{\theta_K}$, and decision $\mathbb{I}\{S(x)\ge \epsilon\}$.

\end{algorithmic}
\end{algorithm}

\subsection{Data Reconstruction Metrics} 
\label{sec:reconstructoin-rewards}
We instantiate reconstruction rewards with lexical similarity metrics $\mathrm{sim}(\hat{y},y^\star)$ that measure the fraction of the reference suffix $y^\star$ reconstructed by a model's completion $\hat{y}$. Since RL learning algorithms suffer from sparse rewards \citep{sutton2018reinforcement}, we propose a variety of surrogate dense rewards that yields sufficient learning signals. All functions $\mathrm{sim}_\star$ share the property that if $y = y^\star$, $\mathrm{sim}_*(y, y^\star) = 1$. Definitions and properties are detailed in \Cref{tab:lexical-rewards}.

\textbf{Token Set Similarity} \cite{morris2025approximating, shao2024scaling}. $\mathrm{sim}_{\text{tok}}(y,y^\star)$ measures the fraction of unique reference tokens in $y^\star$ that appear in $y$, where $T(\cdot)$ denote the set of unique tokens. This metric emphasizes vocabulary recall and penalizes repetition via set overlap.

\textbf{Longest Common Subsequence} \cite{saggion2002meta, lin2004rouge}. $\mathrm{sim}_{\text{lcs}}(y,y^\star)$ uses the length of the longest common subsequence $\mathrm{LCS}(y,y^\star)$, normalized by $\lvert y^\star\rvert$. This rewards correct content in the correct order.

\textbf{N-gram Set Coverage} \cite{hallinan2025surprising}.  $\mathrm{sim}_{\text{ng}}(y,y^\star)$ measures the fraction of reference $n$-grams that appear in $y$, where $\mathcal{N}{[L_{\min},L_{\max}]}(\cdot)$ denotes the set of contiguous $n$-grams with $n\in[L_{\min},L_{\max}]$. Treating $n$-grams as a set to discourage reward hacking via repetition. Following previous works, we fix $L_{\min}=3$ and set $L_{\max} = \mathrm{LCS}(y,y^\star)$.


\begin{table}[t]
\centering
\caption{\textbf{Lexical reconstruction rewards} used for $\mathrm{sim}(\hat{y},y^\star)$.}
\label{tab:lexical-rewards}

\resizebox{\columnwidth}{!}{%
\begin{tabular}{@{}ll@{}}
\toprule
Metric & Definition \\ 
\midrule
Token set similarity &
$\displaystyle
s_{\text{tok}}(\hat{y},y^\star)
=
\frac{\lvert T(\hat{y}) \cap T(y^\star)\rvert}{\lvert T(y^\star)\rvert}
$ \\[0.6ex]
Longest common sequence &
$\displaystyle
s_{\text{lcs}}(\hat{y},y^\star)
=
\frac{\mathrm{LCS}(\hat{y},y^\star)}{\lvert y^\star\rvert}
$ \\[0.6ex]
N-gram set coverage &
$\displaystyle
s_{\text{ng}}(\hat{y},y^\star)
=
\frac{\big\lvert \mathcal{N}_{[L_{\min},L_{\max}]}(\hat{y})\cap \mathcal{N}_{[L_{\min},L_{\max}]}(y^\star)\big\rvert}
{\big\lvert \mathcal{N}_{[L_{\min},L_{\max}]}(y^\star)\big\rvert}
$ \\
\bottomrule
\end{tabular}%
}
\end{table}

\subsection{Contrastive Rewards}
\label{sec:contrastive-rl-objective}

\begin{table*}[t]
\centering
\caption{\textbf{Datasets, models, cutoffs, and maximum sequence lengths for pre-training, post-training, and distillation settings.} Newly constructed MIA datasets are marked with $^\spadesuit$. Cutoff dates marked with * are estimated.}
\label{tab:experiment-setup}

\resizebox{0.7\textwidth}{!}{%
\begin{tabular}{@{\hspace{0.4em}}l l c l c r@{\hspace{0.4em}}}
\toprule
\textbf{Task} & 
\textbf{Dataset} & 
\textbf{Dataset Cutoff} & 
\textbf{Model} & 
\textbf{Model Cutoff} & 
\textbf{Max Tokens} \\
\cmidrule(lr){1-6}

\multirow{3}{*}{Pre-training}
& BookMIA          & 2023 & Llama2-7B     & 2022* & 1024 \\
& WikiMIA$_{2024}$ Hard   & 2024 & Qwen2-7B    & 2023* & 1024 \\
& Dolma3$^\spadesuit$ (Olmo3 Mix)     & 2025 & Olmo3-7B      & 2024 & 2048 \\

\addlinespace[0.8ex]
\cmidrule(lr){1-6}

\multirow{3}{*}{Post-training}
& AIME$^\spadesuit$             & 2024 & Tulu2-7B   & 2023 & 4096 \\
& Olympia Math$^\spadesuit$     & 2025 & Tulu2-7B   & 2023 & 4096 \\
& Tulu3 Mix$^\spadesuit$        & 2024 & Tulu3-8B   & 2024 & 1024 \\

\addlinespace[0.8ex]
\cmidrule(lr){1-6}

\multirow{1}{*}{Distillation}
& S1 / S1.1$^\spadesuit$        & 2025 & Qwen2.5-7B    & 2024* & 2048 \\

\bottomrule
\end{tabular}
}

\end{table*}

Directly maximizing the reconstruction reward in \Cref{eq:Naive-Reward} can introduce overly strong supervision, obscuring the latent membership signal we aim to elicit. To constrain supervision, we use a \emph{contrastive} reward: each rollout $\hat{y}$ is against a pool of reference suffixes $\{g_j\}_{j=0}^K$ containing the true suffix $g_0=y^\star$ and $K$ distractors sampled from other candidates, with the best match as reward. 

Contrastive rewards provide relative supervision (ground truth vs. negative distractors), limiting what the policy can learn beyond what model weights already encode. We propose two contrastive reward variants that form our methods: matching (ADRA) and adaptive matching (ADRA+).

\vspace{.1cm}\noindent\textbf{Matching (ADRA).}
For each candidate $x=(c,y^\star)$, set ${g}_0 := y^\star$ and sample $K$ additional distractor suffixes  $\{g_j\}_{j=1}^K$ i.i.d.\ from $\mathcal{D}_{cand}$.  Given a rollout $\hat{y}\sim p_\theta(\cdot\mid q,c)$, we compute similarities
\[
 \mathrm{sim}(\hat{y}, g_j)\in[0,1],\qquad j=0,\dots,K,
\]
using the reconstruction metrics from \Cref{sec:reconstructoin-rewards}.  
The \emph{matching} reward then scores $\hat{y}$ against all suffixes and takes the maximum:
\begin{equation}
\label{eq:matching-reward}
r_{\text{match}}(\hat{y},x)
\;=\;
\max_{0\le j\le K} \mathrm{sim}(\hat{y}, g_j).
\end{equation}

Because member data is seen during model's training, their rollouts are more likely to match to $g_0$ during matching  $\max_{0\le j\le K} \mathrm{sim}(\hat{y}, g_j)$. We term this method \textsc{ADRA}.

\vspace{.1cm}\noindent\textbf{Adaptive Matching (ADRA+).}
Matching can be further sharpened by incorporating a membership prior $p(x)\in[0,1]$\footnote{We define $p(x)$ such that larger values indicate higher membership likelihood.}, derived from the base model’s log-probabilities (e.g., Min-$K\%$ or Min-$K\%^{++}$ scores). For simplicity, we treat all loss-based variants as $p(x)$ for notation. 

For each $x$ we have the same pool of references $\{g_j\}_{j=0}^K$ and scores $\{\mathrm{sim}_j\}_{j=0}^K$ and define
\begin{align*}
c_{\text{union}} &:= \max_{0\le j\le K} \mathrm{sim}(\hat{y}, g_j), \\
c_{\text{distract}} &:= \max_{1\le j\le K} \mathrm{sim}(\hat{y}, g_j).
\end{align*}
We then sample a Bernoulli switch 
$z \sim \mathrm{Bernoulli}\big(p(x)\big)$
and define the \emph{adaptive matching} reward:
\begin{equation}
\label{eq:adaptive-matching}
r_{\text{adapt}}(\hat{y},x)
=
\begin{cases}
c_{\text{union}}, & z=1,\\
c_{\text{distract}}, & z=0.
\end{cases}
\end{equation}
\noindent where $z=1$ matches among $\{g_0\}\cup\{g_j\}_{j=1}^K$ (true suffix + distractors), and
$z=0$ matches among $\{g_j\}_{j=1}^K$ (distractors only). In practice, for a fixed $(\hat{y},x)$, we compute the expected reward:
\begin{equation}
    \mathbb{E}_z[r_{\text{adapt}}(\hat{y},x)] = p(x)\,c_{\text{union}} + (1-p(x))\,c_{\text{distract}.}
\end{equation}

Intuitively, when $p(x)$ is high (likely member), we often include $g_0$, making it easier for RL to reinforce reconstruction of the true suffix. When $p(x)$ is low, we more often match only against distractors, receiving weaker and noisier supervision. We term this method as \textsc{ADRA+}.

For both variants, the underlying RL algorithm (GRPO) is unchanged: we replace $r(y,x)$ in \eqref{eq:RL-Objective} with either $r_{\text{match}}$ or $r_{\text{adapt}}$.


\section{Experimental Setup}

As shown in \Cref{tab:experiment-setup}, we evaluate membership inference on LLMs spanning three training stages: pre-training, post-training, and distillation. See \Cref{sec:app-training-details} for full details.

\vspace{.1cm}\noindent\textbf{Pre-training.} Following \citep{hallinan2025surprising}, we assess pre-training data detection on \textit{BookMIA} and \textit{WikiMIA24-Hard}. BookMIA \citep{shi2023detecting} contains 512-word snippet sampled from 100 books, and non-members come from books after 2023. WikiMIA$_{\text{2024}}$-Hard \citep{hallinan2025surprising} pairs Wikipedia article versions across date cutoffs and minimizes temporal distribution shift between members and non-members. We use Llama2-7B and Qwen2-7B---their knowledge cutoff aligns well with the dataset cutoffs. In addition, we evaluate the fully-open language model Olmo3 \citep{olmo2025olmo} on their pre-training mix Dolma3.  

\vspace{.1cm}\noindent\textbf{Post-training.} Post-training has become a growing focus of LLMs development 
\citep{guo2025deepseek}. We study two settings: (i) \textit{Controlled contamination:} we simulate test-set contamination by SFT’ing on member examples from AIME and Olympia Math. (ii) \textit{Post-training mixtures:} we evaluate membership on the open Tulu3 post-training mix on Tulu3 model across domain \citep{lambert2024tulu}. 

\vspace{.1cm}\noindent\textbf{Distillation.} Finally, we study data detection under model distillation. Following \citet{muennighoff2025s1}, we construct a dataset of reasoning traces from Deepseek-R1 and Gemini-2.0-Flash. 


\definecolor{hdrblue}{RGB}{224,234,249}

\begin{table*}[t]
\centering
\caption{
\textbf{Pre-training MIA results} on BookMIA, WikiMIA\textsubscript{2024} Hard, and Dolma3 arXiv. \textbf{Bold} denotes the top-2 performance for each dataset. Our active MIA methods (ADRA and ADRA+) consistently outperform passive MIAs.
}
\label{tab:mia-pretrain}
\begingroup
\renewcommand{\arraystretch}{1.25}
\resizebox{0.8\linewidth}{!}{%
\begin{tabular}{@{}l c >{\columncolor{hdrblue}}c >{\columncolor{hdrblue}}c c c c c c c@{}}
\toprule
& & \multicolumn{2}{c}{\textsc{Active MIAs}} &
\multicolumn{6}{c}{\textsc{Passive MIAs}} \\
\cmidrule(lr){3-4} \cmidrule(lr){5-10}
\textbf{Model} & \textbf{Type} 
& \textbf{ADRA+}
& \textbf{ADRA}
& \textbf{N-Sampling}
& \textbf{Loss}
& \textbf{R-Loss}
& \textbf{Zlib}
& \textbf{Min-K\%}
& \textbf{Min-K\%++} \\
\midrule
\sectt{BookMIA}
Llama2-7B  & Orig. & \textbf{78.4} & \textbf{76.2} & 73.6 & 60.6 & N/A & 50.4 & 63.3 & 59.6 \\
Llama2-7B  & Para. & \textbf{74.5} & \textbf{73.0} & 68.7 & 57.2 & N/A & 40.6 & 61.0 & 53.7  \\
\sectt{WikiMIA\textsubscript{2024} Hard}
Qwen2-7B  & Orig. & \textbf{60.6} & \textbf{59.1} & 54.3 & 45.7 & N/A & 44.5 & 45.4 & 50.6 \\
Qwen2-7B  & Para. & \textbf{56.6} & \textbf{58.4} & 55.8 & 47.4 & N/A & 47.3 & 48.3 & 48.5 \\
\sectt{Dolma3 arXiv}
Olmo3-7B-Instruct  & Orig. & \textbf{92.4} & \textbf{91.8} & 81.5 & 71.1 & N/A & 42.3 & 64.4 & 37.4 \\
Olmo3-7B-Instruct  & Para. & \textbf{90.3} & \textbf{84.7} & 70.4 & 69.4 & N/A & 43.9 & 57.5 & 29.9 \\ 
\addlinespace
\bottomrule
\end{tabular}%
}
\endgroup
\end{table*}

\begin{table*}[t]
\centering
\caption{
\textbf{Pre-training member data reconstruction} for original verbatim setting. \textbf{Bold} denotes the best average performance for each dataset.  ADRA+ and ADRA consistently outperform the N-Sampling across all metrics. See \Cref{sec:app-pre-train-reconstruction} for paraphrased setting results.
}
\label{tab:mia-reconstruction-pretraining}
\begingroup
\renewcommand{\sect}[1]{\rowcolor{gray!18}\multicolumn{8}{l}{\textbf{#1}}\\}
\resizebox{0.8\linewidth}{!}{%
\begin{tabular}{@{}lllccccc@{}}
\toprule
\textbf{Model} & \textbf{Type} & \textbf{Method} &
\makecell{\textbf{Budget} \\ (\textbf{$N$})} &
\makecell{\textbf{Lexical Jaccard}\\(Best / Avg)} &
\makecell{\textbf{Lexical LCS}\\ (Best / Avg)} &
\makecell{\textbf{Lexical Coverage}\\ (Best / Avg)} &
\makecell{\textbf{Embedding Cosine}\\ (Best / Avg)} \\
\midrule

\sect{BookMIA}
\addlinespace[0.6ex]
\multirow{4}{*}{Llama2-7B} & \multirow{4}{*}{Orig.}
  & ADRA+ & 32 & \bestavg{21.9}{\textbf{18.5}} & \bestavg{62}{\textbf{54}} & \bestavg{5.7}{\textbf{2.4}} & \bestavg{89.0}{83.9} \\

\cmidrule(lr){3-8}
  &  & ADRA & 32 & \bestavg{18.9}{15.9} & \bestavg{56}{48} & \bestavg{4.2}{2.0} & \bestavg{88.6}{\textbf{84.4}} \\

\cmidrule(lr){3-8}
  &  & N-Sampling & 32 & \bestavg{19.9}{15.6} & \bestavg{57}{46} & \bestavg{4.9}{1.5} & \bestavg{89.6}{82.7} \\
  \addlinespace[1.2ex]




\sect{WikiMIA\textsubscript{2024} Hard}
\addlinespace[0.6ex]
\multirow{4}{*}{Qwen2-7B} & \multirow{4}{*}{Orig.}
  & ADRA+ & 32 & \bestavg{20.1}{14.1} & \bestavg{51}{33} & \bestavg{16.1}{6.0} & \bestavg{88.4}{82.3} \\
\cmidrule(lr){3-8}

  & & ADRA & 32 & \bestavg{17.3}{\textbf{14.8}} & \bestavg{34}{27} & \bestavg{7.0}{4.3} & \bestavg{87.0}{\textbf{84.3}} \\
\cmidrule(lr){3-8}
  &  & N-Sampling & 32 & \bestavg{16.6}{11.4} & \bestavg{52}{\textbf{42}} & \bestavg{16.7}{\textbf{7.6}} & \bestavg{87.9}{81.6} \\
\addlinespace[1.2ex]




\sect{Dolma3 arXiv}
\addlinespace[0.6ex]
\multirow{4}{*}{Olmo3-7B-Instruct} & \multirow{4}{*}{Orig.}
  & ADRA+ & 32 & \bestavg{15.2}{12.6} & \bestavg{69}{\textbf{60}} & \bestavg{9.5}{\textbf{5.0}} & \bestavg{87.5}{\textbf{84.3}} \\
\cmidrule(lr){3-8}

  & & ADRA & 32 & \bestavg{15.9}{\textbf{13.0}} & \bestavg{69}{58} & \bestavg{9.7}{4.9} & \bestavg{87.4}{83.9} \\
\cmidrule(lr){3-8}
  &  & N-Sampling & 32 & \bestavg{15.7}{11.9} & \bestavg{59}{35} & \bestavg{8.1}{2.4} & \bestavg{87.6}{83.4} \\
\addlinespace[1.2ex]



\bottomrule
\end{tabular}
}
\endgroup
\end{table*}

\definecolor{hdrblue}{RGB}{224,234,249}

\begin{table*}[t]
\centering
\caption{
\textbf{Post-training MIA results} on AIME, Olympia Math, and Tulu3 Mix. Our active MIAs (ADRA and ADRA+) outperform calibration-free passive methods and match the calibration-based reference method (R-Loss).
}
\label{tab:post-training-mia}
\begingroup
\renewcommand{\arraystretch}{1.25}
\resizebox{0.8\linewidth}{!}{%
\begin{tabular}{@{}l c >{\columncolor{hdrblue}}c >{\columncolor{hdrblue}}c c c c c c c@{}}
\toprule
& & \multicolumn{2}{c}{\textsc{Active MIAs}} &
\multicolumn{6}{c}{\textsc{Passive MIAs}} \\
\cmidrule(lr){3-4} \cmidrule(lr){5-10}
\textbf{Model} & \textbf{Type} 
& \textbf{ADRA+}
& \textbf{ADRA}
& \textbf{N-Sampling}
& \textbf{Loss}
& \textbf{R-Loss}
& \textbf{Zlib}
& \textbf{Min-K\%}
& \textbf{Min-K\%++} \\
\midrule
\sectt{Olympia Math}
Tulu2-7B  & Orig. & \textbf{70.5}& 67.1 & 59.5 & 53.7 & \textbf{70.3} & 55.6  & 53.2 & 60.2 \\
Tulu2-7B  & Para. & \textbf{68.5} & \textbf{69.4} & 58.2  & 49.6 & 60.1 & 54.9 & 51.1 & 52.9 \\
\sectt{AIME}
Tulu2-7B  & Orig. & \textbf{85.9} & \textbf{85.6} &  72.7 & 71.1 & 67.8 & 61.7 & 75.8 & 78.3 \\
Tulu2-7B  & Para. & \textbf{86.2}  & \textbf{82.2}  & 75.4 & 66.8 & 58.4 & 60.2 & 72.1 & 73.2 \\
\sectt{Tulu3 Mix}
Tulu3-8B  & Aya & \textbf{66.5} & 65.9 & 61.3 & 62.0 & \textbf{68.0 }& 56.4 & 61.6 & 56.7 \\
Tulu3-8B  & Wildchat & \textbf{65.6} & 65.2 & 56.9  & 51.8 & \textbf{66.2}  & 46.7  & 52.8  & 64.3   \\ 
\addlinespace
\bottomrule
\end{tabular}%
}
\endgroup
\end{table*}
\begin{table*}[t]
\centering
\caption{
\textbf{Post-training member data reconstruction} for original verbatim setting.  ADRA+ and ADRA consistently outperform the N-Sampling across both lexical and semantic metrics. See Appendix \Cref{sec:app-post-train-reconstruction} for paraphrased setting results.
}
\label{tab:mia-post-training-data-reconstruction}
\begingroup
\renewcommand{\sect}[1]{\rowcolor{gray!18}\multicolumn{8}{l}{\textbf{#1}}\\}
\resizebox{0.8\linewidth}{!}{%
\begin{tabular}{@{}lllccccc@{}}  
\toprule
\textbf{Model} & \textbf{Type} & \textbf{Method} &
\makecell{\textbf{Budget} \\ (\textbf{$N$})} &
\makecell{\textbf{Lexical Jaccard}\\(Best / Avg)} &
\makecell{\textbf{Lexical LCS}\\ (Best / Avg)} &
\makecell{\textbf{Lexical Coverage}\\ (Best / Avg)} &
\makecell{\textbf{Embedding Cosine}\\ (Best / Avg)} \\
\midrule

\sect{Olympia Math}
\addlinespace[0.6ex]
\multirow{4}{*}{Tulu2-7B} & \multirow{4}{*}{Orig.}
  & ADRA+ & 32 & \bestavg{24.1}{20.7} & \bestavg{68}{59} & \bestavg{15.3}{\textbf{10.9}} & \bestavg{91.5}{88.8} \\

\cmidrule(lr){3-8}
  &  & ADRA & 32 & \bestavg{25.6}{\textbf{23.4}} & \bestavg{79}{\textbf{68}} & \bestavg{12.3}{10.2} & \bestavg{92.8}{\textbf{91.3}} \\

\cmidrule(lr){3-8}
  &  & N-Sampling & 32 & \bestavg{22.4}{15.5} & \bestavg{60}{34} & \bestavg{11.7}{5.0} & \bestavg{92.2}{87.3} \\

\sect{AIME}
\addlinespace[0.6ex]
\multirow{4}{*}{Tulu2-7B} & \multirow{4}{*}{Orig.}
  &  ADRA+ & 32 & \bestavg{22.3}{18.1} & \bestavg{55}{39} & \bestavg{17.2}{11.9} & \bestavg{91.9}{88.9} \\
  \cmidrule(lr){3-8}

  & & ADRA & 32 & \bestavg{21.7}{\textbf{18.5}} & \bestavg{58}{\textbf{43}} & \bestavg{15.5}{\textbf{12.5}} & \bestavg{91.8}{\textbf{89.0}} \\
\cmidrule(lr){3-8}
  &  & N-Sampling & 32 & \bestavg{19.9}{12.9} & \bestavg{50}{27} & \bestavg{14.7}{5.7} & \bestavg{91.8}{85.1} \\
\addlinespace[1.5ex]

\sect{Tulu3 Mix}
\addlinespace[0.6ex]
\multirow{4}{*}{Tulu3-8B} & \multirow{4}{*}{Aya}
  &  ADRA+ & 32 & \bestavg{44.3}{43.9} & \bestavg{14}{14} & \bestavg{47.8}{47.4} & \bestavg{89.2}{88.9} \\
  \cmidrule(lr){3-8}

  & & ADRA & 32 & \bestavg{51.6}{\textbf{51.2}} & \bestavg{16}{\textbf{16}} & \bestavg{53.0}{\textbf{52.5}} & \bestavg{91.5}{\textbf{91.1}} \\
\cmidrule(lr){3-8}
  &  & N-Sampling & 32 & \bestavg{41.6}{26.5} & \bestavg{13}{10} & \bestavg{40.0}{23.9} & \bestavg{90.4}{84.9} \\
\addlinespace[1.5ex]

\multirow{4}{*}{Tulu3-8B} & \multirow{4}{*}{Wildchat}
  &  ADRA+ & 32 & \bestavg{41.4}{\textbf{40.3}} & \bestavg{63}{\textbf{61}} & \bestavg{42.6}{\textbf{40.4}} & \bestavg{92.0}{\textbf{91.4}} \\
  \cmidrule(lr){3-8}

  & & ADRA & 32 & \bestavg{37.5}{34.9} & \bestavg{54}{49} & \bestavg{36.7}{31.7} & \bestavg{91.7}{90.2} \\
\cmidrule(lr){3-8}
  &  & N-Sampling & 32 & \bestavg{35.8}{28.1} & \bestavg{58}{48} & \bestavg{33.7}{24.0} & \bestavg{92.9}{89.5} \\

\bottomrule
\end{tabular}
}
\endgroup
\end{table*}

\subsection{LLM-MIA Dataset Construction}

Most MIA datasets target pre-training and use cutoffs around 2023 \citep{shi2023detecting}. To match frontier LLMs with later cutoffs and substantially more post-training data, we develop 6 new LLM-MIA datasets.

\textbf{Dolma3 / Tulu3 Mix}: Fully open models enable controlled membership evaluation. We construct pre-training and post-training MIA datasets from the Dolma3 pre-training mix \citep{olmo2025olmo} and Tulu3 SFT mix \cite{lambert2024tulu}. For Dolma3, we choose arXiv documents as members, and sample non-members from later-dated arXiv papers beyond the mix cutoff. For Tulu3, we focus on Aya \citep{singh2024aya} and WildChat \citep{zhao2024wildchat}; members are the subset included in the Tulu3 mix while non-members are excluded in the mix.

\textbf{AIME / Olympiad Math:} We construct controlled-contamination math MIA datasets by defining members as the data used for SFT and non-members as held-out data. For AIME, we collect problems from 2021--2025, using 2021--2024 as members and 2025 as non-members. For Olympia Math \citep{olympiads_ref}, we split problems into member and non-member under random partitions.

\textbf{S1 / S1.1 Distillation}: Following \citet{muennighoff2025s1}, we use reasoning traces from two teacher models, Gemini-2.0-Flash (S1) and DeepSeek-R1 (S1.1). We treat either group as members depending on the experiment.

Overall, we evaluate on 8 MIA datasets. Additional to the verbatim (original) setting, we evaluate the \textit{paraphrased} setting by paraphrasing the constructed datasets with Gemini-2.5-Flash. See \Cref{sec:app-dataset-curation} for full dataset construction details.

\subsection{MIA Baselines \& Evaluation}

Following \citep{duan2024membership}, we consider five loss-based baselines, each described in \Cref{sec:background}. We also include the reconstruction-based baseline from \citep{hallinan2025surprising}: given prefix, sample $N$ continuations and score each by n-gram overlap metrics to the ground truth suffix. We generalize \citep{hallinan2025surprising}'s method to \textsc{N-Sampling}: score each continuation with a suite of similarity metrics, and aggregate scores using either the Average-of-N (AoN) or Best-of-N (BoN). We use lexical metrics in \Cref{tab:lexical-rewards} and include embedding metrics and length-normalized variants. 

We report AUROC for MIA performance, following \citep{shi2023detecting, duan2024membership, hallinan2025surprising}, and evaluate reconstruction using the same lexical and embedding similarity metrics. 

For full metric definitions and evaluation details see \Cref{sec:app-evaluation-details}.

\section{Results}
\subsection{Pre-training}


\Cref{tab:mia-pretrain} reports AUROC (\%) for pre-training MIAs, comparing \Method/\MethodPlus~to N-Sampling and 4 loss-based baselines. We find that \Method~and \MethodPlus~achieve significant improvements over current methods. In the original setting, \MethodPlus~on average outperforms the best loss-based MIAs by {15.1\%, 10.0\%, 21.3\%} in BookMIA, WikiMIA\textsubscript{2024} Hard, and Dolma3 arXiv respectively; \MethodPlus~improves over N-sampling consistently by at least 5\% in all benchmarks. 
In the paraphrased setting, \MethodPlus~is also the best-performing approach across all datasets. 

\Cref{tab:mia-reconstruction-pretraining} shows the member reconstruction quality on N-Sampling, \Method, \MethodPlus~in the original setting. On average, we find that \Method~and \MethodPlus~consistently improve data reconstruction, particularly in the lexical Jaccard and embedding cosine. On WikiMIA\textsubscript{2024} Hard, RL increases  Jaccard and embedding metrics but reduces LCS and n-gram coverage, suggesting a metric trade-off caused by reward dominance. Results on paraphrased setting are nearly identical and can be found in \Cref{sec:app-pre-train-reconstruction}. Overall, RL-induced data reconstructions yield more faithful extracts and correspondingly stronger membership signals.

\subsection{Post-training}

\Cref{tab:post-training-mia} reports AUROC (\%) under two post-training regimes: (i) controlled contamination and (ii) post-training mixture.
For AIME and Olympia Math, we simulate contamination by finetuning Tulu2-7B on held-out member examples while mixing in Tulu2-Mix, with a 10\% contamination rate.
For the mixture setting, we evaluate Tulu3-8B on Aya and WildChat from the Tulu3 Mix. For R-Loss, we calibrate with backbone-matched references: Llama2-7B for Tulu2-7B and Llama3-8B for Tulu3-8B. 

On Olympia Math and AIME, \MethodPlus~achieves the best AUROC in both original and paraphrased settings, and \Method~is consistently second.
Relative to N-Sampling, \MethodPlus~improves AUROC by 11.0\% on Olympia Math, from 59.5\% to 70.5\%, and by 13.2\% on AIME, from 72.7\% to 85.9\%, in the original setting, with similar gains under paraphrasing.
On Tulu3 SFT Mix, \MethodPlus~equals R-Loss on WildChat (65.6\%) AUROC and matches Aya (66.5\% vs. 68.0\%), while outperforming the remaining baselines.
Unlike R-Loss, our attack does not require an additional calibration model.

\Cref{tab:mia-post-training-data-reconstruction} shows post-training member reconstruction quality. Compared to pre-training in \Cref{tab:mia-reconstruction-pretraining}, on average our method achieves higher reconstruction on all metrics than N-Sampling. The gains from RL finetuning are more pronounced over sampling: on Olympia Math for example, average Jaccard improves by 7.9, LCS by 34, coverage by 5.9, and embedding cosine by 4.9. Overall, post-training data appear more extractable than pre-training, consistent with stronger memorization from finetuning late in the training pipeline.


\subsection{Distillation}
\definecolor{hdrblue}{RGB}{224,234,249}
\newcommand{\secttt}[1]{\rowcolor{gray!18}\multicolumn{7}{l}{\textbf{#1}}\\}

\begin{table}[t]
\centering
\caption{
\textbf{Distillation MIA results.} \Method~outperforms all passive MIAs on both distillation datasets (S1.1 and S1).
}
\label{tab:mia-distillation}
\begingroup
\renewcommand{\arraystretch}{1.25}
\resizebox{1.0\linewidth}{!}{%
\begin{tabular}{@{}>{\columncolor{hdrblue}}c c c c c c c@{}}
\toprule
\multicolumn{1}{c}{\textsc{Active MIA}} &
\multicolumn{6}{c}{\textsc{Passive MIAs}} \\
\cmidrule(lr){1-1} \cmidrule(lr){2-7}
\multicolumn{1}{c}{\cellcolor{hdrblue}\textbf{ADRA}}
& \textbf{N-Sampling}
& \textbf{Loss}
& \textbf{R-Loss}
& \textbf{ZLib}
& \textbf{Min-K\%}
& \textbf{Min-K\%++} \\
\midrule

\secttt{S1.1 Distillation: Deepseek-R1}
 \multicolumn{1}{c}{\cellcolor{hdrblue}\textbf{98.4}} & 90.4  & 55.1 & 70.0 & 52.6 & 55.7 & 70.6 \\

\secttt{S1 Distillation: Gemini-2.0-flash}
 \multicolumn{1}{c}{\cellcolor{hdrblue}\textbf{85.2}} & 80.8  & 69.8 & 79.3 & 74.3 & 71.9 & 81.7 \\
\bottomrule
\end{tabular}%
}
\endgroup
\end{table}

\begin{table}[t]
\centering
\caption{
\textbf{Distillation member data reconstruction.} ADRA achieves significantly better reconstruction across all metrics.
}
\label{tab:mia-distillation-data-reconstruction}
\begingroup
\renewcommand{\sect}[1]{\rowcolor{gray!18}\multicolumn{5}{l}{\textbf{#1}}\\}
\resizebox{\linewidth}{!}{%
\begin{tabular}{@{}lcccc@{}}
\toprule
\textbf{Method} &
\makecell{\textbf{Lexical Jaccard}\\(Best / Avg)} &
\makecell{\textbf{Lexical LCS}\\ (Best / Avg)} &
\makecell{\textbf{Lexical Coverage}\\ (Best / Avg)} &
\makecell{\textbf{Embedding Cosine}\\ (Best / Avg)} \\
\midrule

\sect{S1.1 Distillation: Deepseek-R1}
\addlinespace[0.6ex]
ADRA        & \bestavg{45.4}{\textbf{44.5}} & \bestavg{145}{\textbf{143}} & \bestavg{48.6}{\textbf{47.7}} & \bestavg{97.6}{\textbf{97.5}} \\
N-Sampling  & \bestavg{32.7}{25.5}          & \bestavg{111}{82}          & \bestavg{34.9}{22.3}          & \bestavg{97.4}{95.6}          \\

\sect{S1 Distillation: Gemini-2.0-Flash}
\addlinespace[0.6ex]
ADRA        & \bestavg{31.0}{\textbf{27.4}} & \bestavg{117}{\textbf{106}} & \bestavg{32.8}{\textbf{28.3}} & \bestavg{96.4}{\textbf{95.6}} \\
N-Sampling  & \bestavg{27.8}{21.2}          & \bestavg{111}{80}          & \bestavg{27.4}{17.5}          & \bestavg{96.2}{93.7}          \\

\bottomrule
\end{tabular}
}
\endgroup
\end{table}

\Cref{tab:mia-distillation} reports AUROC (\%) in a distillation setting where a student (Qwen2.5-7B-Instruct) is trained for one epoch on member reasoning traces produced by a teacher (Deepseek-R1 or Gemini-2.0-Flash), following \citet{muennighoff2025s1}.  We do not mix additional data, so the distilled traces constitute the entire training data. In this setup, member and non-member examples share the \emph{same prefix} and differ only in the generated suffix, making \MethodPlus~not directly applicable (see \Cref{sec:app-contrastive-reward-details}). We therefore use \Method~only with a generation budget of $N=16$. 

Notably, \Method~yields near-perfect membership inference under DeepSeek-R1 distillation, reaching 97.0\%, and achieves 86.9\% under Gemini distillation. It improves over N-Sampling by 8.0\% and 4.4\%, and over the strongest loss-based baseline Min-K\%++ by 27.8\% and 3.5\%, respectively. As reported in \Cref{tab:mia-distillation-data-reconstruction}, \Method~also substantially improves reconstruction over N-Sampling in both settings; for example, average Jaccard increases by 19.0 and 6.2, and coverage by 25.4 and 10.8, under DeepSeek-R1 and Gemini-2.0-Flash distillation, respectively.


\section{Ablations \& Analysis}

\begin{figure}[!t]
    \centering
    \resizebox{0.9\linewidth}{!}{%
    \includegraphics{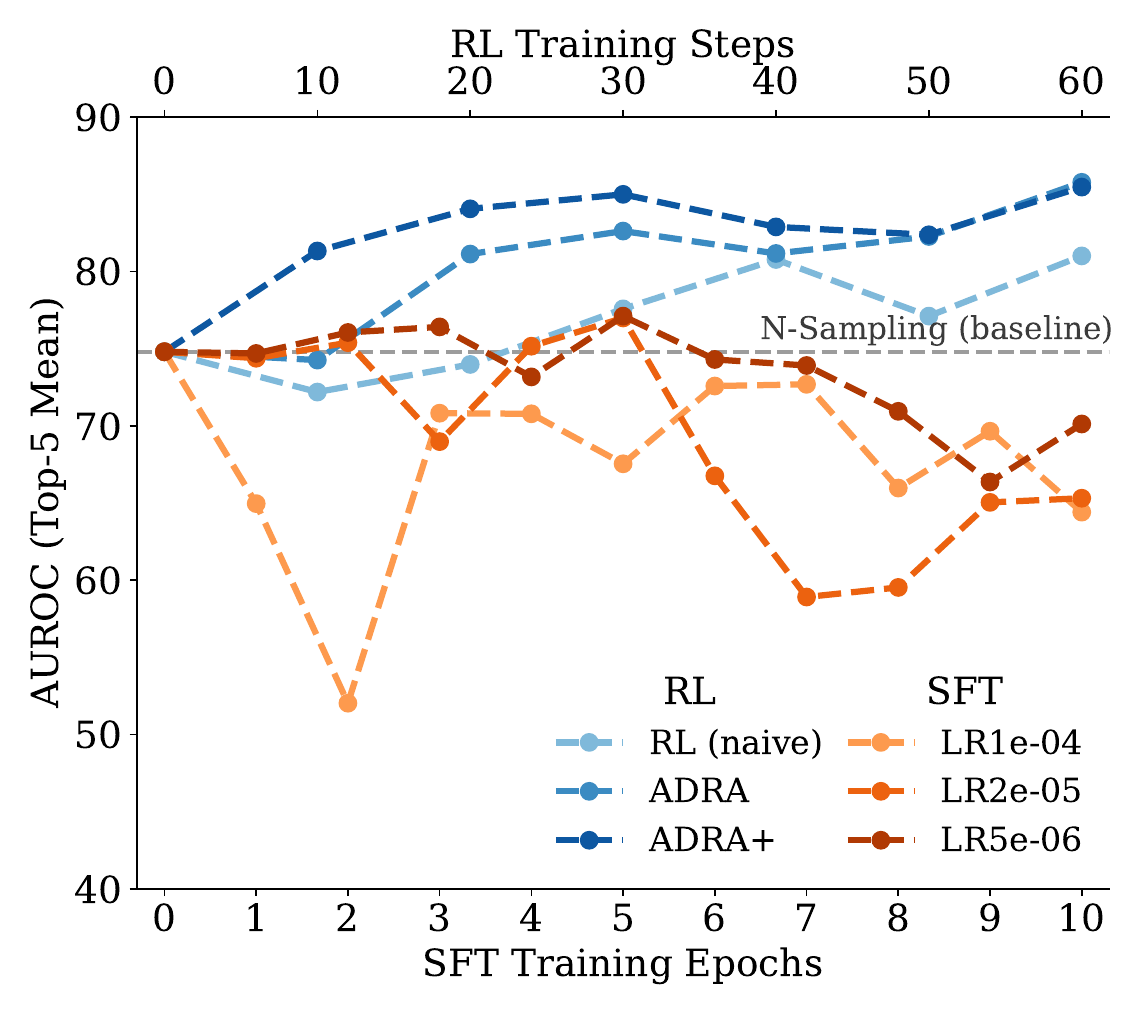} 
    }
    \caption{\textbf{Performance comparison between RL and SFT.} As RL training continues, AUROC improves, whereas SFT decreases. \Method~and \MethodPlus~meaningfully improves over naive RL. }
    \label{fig:ablations-experiment}
\end{figure}

\paragraph{RL vs. SFT.} We perform ablation studies by comparing our method with supervised fine-tuning, across 3 learning rate, and removing our proposed contrastive RL-objectives. We run on AIME on first seed and report top-5 average AUROC scores evaluated from our suite of similarity metrics. As shown in \Cref{fig:ablations-experiment}, RL-based optimization and contrastive objectives are critical to gains in MIA performance. 


\vspace{.1cm}\noindent\textbf{Model Size.} We scale to Tulu2-13B on AIME (\Cref{tab:mia-ablation-13b}). Results mirrors Tulu2-7B, with \Method~and \MethodPlus~outperforming existing MIAs.



\vspace{.1cm}\noindent\textbf{Reward Type.} We compare our lexical reconstruction rewards with two model-based alternatives under non-contrastive RL: embedding (Qwen3-8B-Embedding) and LLM-as-judge (Qwen3-32B). Shown in \Cref{tab:ablation-reward-type}, both model-based variants improve over baseline but achieve lower performance than lexical rewards. Lexical rewards are directly verifiable, while model-based rewards are more prone to reward hacking. See \Cref{sec:app-ablation-details} for implementation details.

\definecolor{hdrblue}{RGB}{224,234,249}
\begin{table}[t]
\centering
\caption{
\textbf{Tulu2-13B MIA results} on AIME. 
}
\label{tab:mia-ablation-13b}
\begingroup
\renewcommand{\arraystretch}{1.25}
\resizebox{1.0\linewidth}{!}{%
\begin{tabular}{@{}>{\columncolor{hdrblue}}c >{\columncolor{hdrblue}}c c c c c c c@{}}
\toprule
\multicolumn{2}{c}{\textsc{Active MIAs}} &
\multicolumn{6}{c}{\textsc{Passive MIAs}} \\
\cmidrule(lr){1-2} \cmidrule(lr){3-8}
\multicolumn{1}{c}{\cellcolor{hdrblue}\textbf{\MethodPlus}}
& \multicolumn{1}{c}{\cellcolor{hdrblue}\textbf{\Method}}
& \textbf{N-Sampling}
& \textbf{Loss}
& \textbf{R-Loss}
& \textbf{ZLib}
& \textbf{Min-K\%}
& \textbf{Min-K\%++} \\
\midrule
 \multicolumn{1}{c}{\cellcolor{hdrblue}\textbf{89.5}} & \multicolumn{1}{c}{\cellcolor{hdrblue}\textbf{84.4}} & 72.7 & 70.7 & 63.3 & 60.6 & 77.7 & 82.8 \\
\bottomrule
\end{tabular}%
}
\endgroup
\end{table}




\section{Related Work}
\label{}
\paragraph{Membership Inference Attacks (MIA).} Given a target model and candidate data, MIA determines whether that data is in the model's training set \citep{shokri2017membership}. In language modeling, most prior work uses the model's log likelihoods on the candidate sequence as a proxy for membership. \citet{yeom2018privacy} directly threshold loss, while \citet{carlini2021extracting} calibrate the target’s loss with a reference model. \citet{shi2023detecting} propose Min-$K\%$ to score only the lowest $k\%$ log-probabilities, and \citet{zhang2024min} normalize {Min-$K\%$ by the expected token log-probabilities over the vocabulary. Other works combine loss with zlib compression \citep{carlini2021extracting}.

A smaller line of works leverages model's generations. \citet{duarte2024cop} leverage models' tendency to distinguish verbatim training data from paraphrased counterparts. Closest to our work is \citep{hallinan2025surprising}, which finds that by generating multiple samples given 50\% of the prefixes, simple n-gram metrics on these generations approach the performance of loss-based MIAs. Despite using generation, these methods remain passive: they elicit one-shot continuations under fixed prompts/prefixes and then score those outputs post hoc. One prior work \citep{tramèr2022truthserumpoisoningmachine} proposes a class of training-based membership inference attacks that poison training data to increase likelihood of extraction of other non-poisoned data. 

In contrast, our approach to MIA actively incentivizes LLMs to reconstruct their own training data. By maximizing a reconstruction-aligned reward, we trained a generative policy that continually up weights member sequences and elicits memorization. 



\paragraph{Training Data Detection in LLMs.} Reliably detecting training data is a key challenge: training corpora span trillions of tokens and, for many frontier models, the underlying datasets are undisclosed \citep{touvron2023llama, balloccu2024leak}. \citet{maini2024llm, duan2024membership, das2025blind} have shown that the efficacy of MIAs is confounded by the distribution shifts between member and non-members. Rigorously proving data membership requires low false-positive rate \citet{carlini2022membership, zhang2024membership}, which is difficult to evaluate on close-data models. Instead of inferring membership about individual data, \cite{maini2024llm} proposes dataset inference for LLMs to detect datasets used for training. In parallel, \citep{jacovi2023stop, elazar2023s} find test-set contamination to be a growing concern in LLMs evaluation \citep{jacovi2023stop, elazar2023s}. Paraphrases, translations, and synthetic restatements can inflate benchmark scores while evading detection \citep{yang2023rethinking, dekoninck2024evading}. \citet{sainz2023nlp, singh2024evaluation, liu2025language} argues that data membership for LLMs should be defined behaviorally rather than exact match. Our work explores both \textit{strict} membership—verbatim inclusion—and \textit{soft} membership, which counts semantically or functionally equivalent variants such as paraphrases as members. Under this formulation, false positives are less concerning as semantically equivalent data can constitute effective contamination. 

\begin{table}[t]
\centering
\caption{
\textbf{Reward type comparison.} Lexical rewards outperform generative-based alternatives (Embedding and LLM-as-Judge).
}
\label{tab:ablation-reward-type}
\resizebox{\linewidth}{!}{%
\begin{tabular}{@{}lcccc@{}}
\toprule
\textbf{Method} & Lexical & Embedding & LLM-as-Judge & No-Training \\
\midrule
\textbf{AUROC} (Top-5 Mean) & \textbf{81.3} & 76.9 & 78.8 & 74.8 \\
\bottomrule
\end{tabular}%
}
\end{table}
\vspace{-0.08in}
\paragraph{Training Data Extraction and Memorization.} Modern language models are known to memorize and regurgitate training data, which includes harmful and copyrighted information \citep{carlini2021extracting,biderman2023emergentpredictablememorizationlarge,nasr2023scalableextractiontrainingdata,zhang2023counterfactualmemorizationneurallanguage,cooper2025filescomputercopyrightmemorization}. Pretraining dataset size often far exceeds model size, making data extraction difficult \citep{morris2025languagemodelsmemorize}. Models are also often intentionally aligned to make extraction more difficult after the alignment stage. Even with these theoretical and empirical safeguards in place, models still exhibit memorization of data after finetuning and post-training \citep{barbero2025extractingalignmentdataopen, ahmed2026extracting}. Data extraction commonly works by \textit{discoverable extraction}: splitting a data into a prefix and a suffix, prompting the LLM with the prefix, and evaluating similarity between sampled generations and the suffix \citep{carlini2021extracting, hayes2025measuring}. Our work improves extraction by fine-tuning the target model with on-policy RL, actively eliciting latent memorization through weight updates.



\section{Conclusion} We propose Active Data Reconstruction Attack (ADRA) that leverages reinforcement learning for data detection and reconstruction. ADRA is the first active MIA, consistently improving over passives MIAs across all stages of model training. Results suggest that model weights encode more about training data than previous methods reveal. 


\newpage
\section*{Impact Statement}

This paper investigates whether training data can be detected and reconstructed from language model weights through membership inference and active reconstruction methods. Our goal is to understand the privacy risks inherent in current language models across their lifecycle, from pre-training through post-training. The purpose of this work is to evaluate the limits of existing privacy protections and to motivate the development of more robust defenses against unintended data leakage.

\section*{Acknowledgment}
Junjie Oscar Yin thanks Rulin Shao, Zhiyuan Zeng, Hamish Ivison, Weijia Shi for the helpful comments and discussions, Lea Li, Jacqueline He, Stella Li for the support. This research was developed with funding from the Defense Advanced Research Projects Agency's (DARPA) SciFy program (Agreement No. HR00112520300) and NSF Grant No. IIS2142739. The views expressed are those of the author and do not reflect the official policy or position of the Department of Defense or the U.S. Government.


\bibliography{icml2026}
\bibliographystyle{icml2026}

\newpage
\appendix
\onecolumn


\section{Experimental Details}
\label{sec:app-training-details}
We provide implementation details for replicating our experimental results, including prefix construction and hyperparameters for each training stage. All trainings are done in a single node consisting of 8 H200s.

\subsection{Pre-training}
Following \citet{hallinan2025surprising}, we use approximately 50\% of the candidate data as the prefix and reconstruct the remaining held-out suffix. We limit generation length to half the maximum token count of the dataset. We list the prompts we use before the candidate prefix in \Cref{tab:pre-training-prompts}. 

\definecolor{SkinColor}{RGB}{253, 247, 244}
\definecolor{SkyBlue}{RGB}{103, 143, 195}

\begin{table}[t]
\centering
\caption{Prompts used for pre-training experiments across datasets. The \texttt{\{prefix\}} placeholder indicates where the input text is inserted.}
\label{tab:pre-training-prompts}
\small
\begin{tabular}{p{0.22\textwidth}p{0.73\textwidth}}
\toprule
\textbf{Dataset} & \textbf{Prompt Template} \\
\midrule
\textbf{BookMIA} & 
\begin{tcolorbox}[colframe=SkyBlue, colback=blue!2, boxsep=2pt, left=3pt, right=3pt, top=3pt, bottom=3pt, arc=2pt]
\small
You will receive a prefix from a passage and be asked to complete it based on the text of a famous work. Provide only the continuation for the last given prefix without any extra commentary, formatting, or additional text. 

Complete the prefix: \texttt{\{prefix\}}
\end{tcolorbox} \\
\midrule
\textbf{WikiMIA\textsubscript{2024} Hard} & 
\begin{tcolorbox}[colframe=SkyBlue, colback=blue!2, boxsep=2pt, left=3pt, right=3pt, top=3pt, bottom=3pt, arc=2pt]
\small
Continue the generation as closely to verbatim as possible. 

\texttt{\{prefix\}}
\end{tcolorbox} \\
\midrule
\textbf{Dolma3 arXiv} & 
\begin{tcolorbox}[colframe=SkyBlue, colback=blue!2, boxsep=2pt, left=3pt, right=3pt, top=3pt, bottom=3pt, arc=2pt]
\small
Continue the generation as closely to verbatim as possible. 

\texttt{\{prefix\}}
\end{tcolorbox} \\
\bottomrule
\end{tabular}
\end{table}

\subsection{Post-training}
\paragraph{Controlled contamination.} To simulate test-set contamination, we fine-tune on member examples from AIME and Olympia Math. We leverage the natural problem-solution structure in both math datasets by including the full problem plus 25\% of the solution as the prefix. To simulate practical contamination scenarios, we also mix member examples with other SFT data at a fixed 10\% contamination rate using the Tulu2 SFT mix \citep{ivison2023camels}. \Cref{tab:appendix-sft-hyperparameters} details our fine-tuning hyperparameters.

\paragraph{Post-training mixtures.} We leverage dataset structure to determine the amount of prefix. Both Aya and WildChat are instruction-tuning datasets containing user requests and assistant responses, and our curated MIA datasets only contain single-turn conversations. We include the full user request plus 25\% of the assistant response as the prefix. 

\begin{table}[t]
\centering
\small
\setlength{\tabcolsep}{6pt}
\caption{Supervised fine-tuning hyperparameters for controlled contamination setting.}
\label{tab:appendix-sft-hyperparameters}
\resizebox{\linewidth}{!}{%
\begin{tabular}{lccccccc}
\toprule
\textbf{Dataset} & \textbf{Learning rate} & \textbf{Eff. batch size} & \textbf{LR Schedule} &
\textbf{Precision} & \textbf{Epochs} &
\textbf{Warmup ratio} & \textbf{Max seq. len.} \\
\midrule
AIME      & 2e$-$5 & 64 & Linear &
\multirow{2}{*}{BF16} & \multirow{2}{*}{1} &
\multirow{2}{*}{0.00} & \multirow{2}{*}{4096} \\
\cmidrule(lr){1-4}
Olympia Math       & 5e$-$6 & 128 & Constant &  &  &  &  \\
\bottomrule
\end{tabular}
}
\end{table}

\subsection{Distillation}
S1 and S1.1 consist primarily of problem-solution pairs. Since problems and solutions have comparable lengths, we include only the problem as the prefix. Following \citet{muennighoff2025s1}, we perform model distillation on Qwen2.5-7B-Instruct by fine-tuning on the entire S1 or S1.1 dataset without additional data mixing. \Cref{tab:appendix-distillation-hyperparameters} details our fine-tuning hyperparameters. Qualitatively, we find that Qwen2.5-7B-Instruct output resembles that of Deepseek-R1 more closely, so a smaller learning rate is used relative to distilling from Gemini-2.0-Flash.

\begin{table}[t]
\centering
\small
\setlength{\tabcolsep}{6pt}
\caption{Supervised fine-tuning hyperparameters for model distillation setting.}
\label{tab:appendix-distillation-hyperparameters}
\resizebox{\linewidth}{!}{%
\begin{tabular}{lccccccc}
\toprule
\textbf{Dataset} & \textbf{Learning rate} & \textbf{Eff. batch size} & \textbf{LR Schedule} &
\textbf{Precision} & \textbf{Epochs} &
\textbf{Warmup ratio} & \textbf{Max seq. len.} \\
\midrule
S1 - Gemini-2.0-Flash     & 2e$-$5 &
\multirow{2}{*}{128} & \multirow{2}{*}{Linear} &
\multirow{2}{*}{BF16} & \multirow{2}{*}{1} &
\multirow{2}{*}{0.00} & \multirow{2}{*}{8192} \\
\cmidrule(lr){1-2}
S1.1 - Deepseek-R1       & 5e$-$6 &  &  &  &  &  &  \\
\bottomrule
\end{tabular}
}
\end{table}

\section{MIA Dataset Construction}
\label{sec:app-dataset-curation}

\definecolor{SkinColor}{RGB}{253, 247, 244}
\definecolor{SkyBlue}{RGB}{103, 143, 195}
\definecolor{DarkSkyBlue}{RGB}{70, 100, 140}

\begin{figure}[t]
\centering
\begin{tcolorbox}[
    colframe=SkyBlue,
    colback=SkinColor,
    coltitle=white,
    colbacktitle=DarkSkyBlue,
    title=\textbf{Paraphrasing Prompt Template},
    fonttitle=\bfseries,
    boxsep=5pt,
    left=8pt,
    right=8pt,
    top=5pt,
    bottom=5pt,
    arc=3pt,
    width=\textwidth
]
\small
You will be given an ORIGINAL block that may contain natural-language, math, code, or other texts. Your job is to rephrase the natural-language content while mostly PRESERVING NON-NATURAL-LANGUAGE CONTENT.

\begin{enumerate}
\item Please ensure the paraphrased output have similar length to the original text ($\pm$15\%).

\item When encountering mathematical formulas, if necessary, you can replace variable names. For example, you can replace `x' with `y' or `a'.

\item No extra commentary: Do not add explanations, apologies, or any other text besides the required output.

\item Output directly the paraphrased output after ``\texttt{rewrite\_output:}''.
\end{enumerate}

\vspace{0.5em}
\texttt{ORIGINAL: \{input\}}
\end{tcolorbox}
\caption{Prompts used to paraphrase datasets. The \texttt{\{input\}} placeholder indicates where the input text is inserted.}
\label{fig:paraphrase_prompt}
\end{figure}
We provide details on our MIA dataset curation process. Due to computational constraints, we sample subsets from each constructed dataset and average results across multiple random seeds.

In addition to the original (verbatim) setting, we also evaluate a paraphrased setting where member and non-member examples are rewritten while preserving their semantic content. Specifically, we use the \textsc{Gemini-2.5-Flash-v3.0} model API via LiteLLM to generate paraphrased instances. For experiments evaluating existing models, we use paraphrased examples as the evaluation set. For experiments involving fine-tuning, we use paraphrased examples as input to the training set and original examples as the evaluation set. We list the prompts used for each dataset in \Cref{fig:paraphrase_prompt}.

\subsection{Dolma3 Pre-training Mix}
For the pre-training setting, we collect arXiv documents from the Dolma3 pre-training data mix. These documents originate from the Proof-Pile-2 dataset \citep{azerbayev2023llemma} and have a cutoff date of April 2023. We sub-sample 3,000 documents to form our member set. To construct non-members, we collect 1,000 arXiv papers released after 2025 and verify using OlmoTrace that they are not present in the pre-training data. For each seed, we sample 128 examples (64 members, 64 non-members).

\subsection{Tulu3 SFT Mix}
For the post-training mix setting, we collect two datasets from the Tulu3 SFT Mix: Aya and WildChat. Aya is a multilingual instruction-tuning dataset containing human-annotated prompt-completion pairs \citep{singh2024aya}, while WildChat contains real-world user interactions with ChatGPT \citep{zhao2024wildchat}. For both datasets, members are sampled from data included in the Tulu3 SFT Mix, and non-members are data from the original datasets that were excluded from the mix. To control for length variation between members and non-members, we limit both user and assistant turns to a maximum of 1,024 tokens. For each seed, we sample 128 examples (64 members, 64 non-members).

\subsection{AIME / Olympia Math}
For the controlled-contamination setting, we collect challenging math datasets: AIME and Olympia Math. For AIME, we collect problems and solutions from 2021--2025, using 2021--2024 as members and 2025 as non-members. Because each year contains only 30 problems, our member set contains 120 examples while the non-member set contains 30. For each seed, we sample 32 examples (16 members, 16 non-members). For Olympia Math \citep{olympiads_ref}, because the dataset was released in 2025, we split problems into members and non-members using random partitions. For each seed, we sample 64 examples (32 members, 32 non-members).

\subsection{S1 / S1.1}
For the distillation setting, we follow \citet{muennighoff2025s1} and collect reasoning trace datasets from two teacher models: Gemini-2.0-Flash (S1) and DeepSeek-R1 (S1.1). Depending on the experiment, we treat either group as members while the other serves as non-members. For each seed, we sample 256 examples (128 members, 128 non-members).


\section{Active Data Reconstruction Attack Details}
\label{sec:app-method-details}

Below we provide more details on implementation of our method Active Data Reconstruction Attck.

\subsection{GRPO Hyperparamters}

We implement our method ADRA using \textsc{verl} and the GRPO training hyperparameters are listed in \Cref{tab:grpo-hparams}.

\begin{table}[t]
\centering
\caption{GRPO hyperparameters used with \texttt{verl}. Only core settings are shown; batch size, rollouts, sequence lengths, and training epochs vary by dataset.}
\label{tab:grpo-hparams}
\begin{tabular}{@{}ll@{}}
\toprule
\textbf{Hyperparameters} & \textbf{Setting} \\
\midrule
Advantage estimator & GRPO \\
Learning rate (actor) & $5e-05$ \\
Normalize advantages (GRPO) & True \\
Use KL \emph{loss} (not reward penalty) & True \\
KL loss coef. & 0.005 \\
Entropy coefficient & 0 \\
Generation temperature & 1.0 \\
Generation top-$p$ & 0.95 \\
Generation top-$k$ & 50 \\
LoRA rank / $\alpha$ & 64 / 128 \\
Critic warmup & 0 \\
\bottomrule
\end{tabular}%
\end{table}


\subsection{Reconstruction Metrics Details} 

We propose three lexical reconstruction metrics from which our policy attempts to maximize, as detailed in \Cref{tab:lexical-rewards}. In practice, we experiment with two aggregated variants: (1) \textsc{Trio}, which averages token set similarity, longest common subsequence, and N-gram set coverage; and (2) \textsc{N-Gram}, which uses N-gram set coverage alone.

To constrain generation length and prevent reward hacking through excessive output, we apply a length penalty to both metrics. The penalty operates on a threshold-based ratio system: candidate outputs are penalized if their word count falls outside the range $[\ell_{\text{ref}}/\tau, \ell_{\text{ref}} \times \tau]$, where $\ell_{\text{ref}}$ is the reference length and $\tau$ is the threshold parameter. We set $\tau = 1.50$ for both \textsc{Trio} and \textsc{N-Gram}, allowing candidates to be as short as 67\% or as long as 150\% of the reference length without penalty. Outside this range, a linear penalty proportional to the length deviation is applied, multiplying the base similarity score by $\min(1.0, \min(\ell_{\text{cand}}/\ell_{\text{ref}}, \ell_{\text{ref}}/\ell_{\text{cand}}))$.

We provide dataset-specific hyperparameters for these metrics in \Cref{tab:grpo-dataset-hparams}.

\begin{table}[t]
\centering
\caption{Dataset-specific hyperparameters for ADRA training.}
\label{tab:grpo-dataset-hparams}
\resizebox{\linewidth}{!}{%
\begin{tabular}{@{}lccccccc@{}}
\toprule
\textbf{Dataset} & \textbf{Batch size} & \textbf{Rollouts} & \textbf{Max prompt / response tokens} & \textbf{Steps} & \textbf{Reconstruction Metrics} & \textbf{Distractors $K$} & \textbf{Length penalty} \\
\midrule
BookMIA & 64 & 32 & 1028 / 1028 & 30 & \textsc{Trio}, \textsc{N-Gram} & 7 & 1.5 \\
WikiMIA\textsubscript{2024} Hard & 64 & 32 & 1028 / 1028 & 30 & \textsc{Trio}, \textsc{N-Gram} & 7 & 1.5 \\
Dolma3 arXiv & 64 & 32 & 1028 / 1028 & 30 & \textsc{Trio}, \textsc{N-Gram} & 7 & 1.5 \\
AIME & 64 & 32 & 2048 / 2048 & 100 & \textsc{Trio}, \textsc{N-Gram} & 7 & 1.5 \\
OlympiaMath & 64 & 32 & 2048 / 2048 & 100 & \textsc{Trio}, \textsc{N-Gram} & 7 & 2.0 / 1.5 \\
Tulu3 Aya / Wildchat & 64 & 32 & 1024 / 512 & 200 & \textsc{Trio}, \textsc{N-Gram} & 7 & 1.5 \\
S1 / S1.1 & 64 & 16 & 1028 / 4096 & 200 & \textsc{Trio}, \textsc{N-Gram} & N/A & 1.25 \\
\bottomrule
\end{tabular}%
}
\end{table}

\subsection{Contrastive Reward Details}
\label{sec:app-contrastive-reward-details}
We use $K=7$ distractors, so each generation is scored against 8 candidate suffixes: 1 ground truth plus 7 distractors. Distractors are sampled i.i.d. from the evaluation set. 

For WikiMIA$_{\text{2024}}$ Hard, which contains paired Wikipedia articles from different date cutoffs (one member, one non-member), we include the paired article as one distractor in addition to $K-1$ randomly sampled distractors.

In the distillation setting, member and non-member pairs share identical prefixes. We treat each such pair as a single example and maximize the reconstruction score for whichever suffix (member or non-member) achieves the higher lexical metric.





\section{Evaluation Details}
\label{sec:app-evaluation-details}

\subsection{Evaluation Metrics}

We evaluate reconstruction quality using a comprehensive suite of lexical and semantic similarity metrics. For each input, we sample $N$ candidate completions and report both the \textbf{average-of-$N$} (mean score across all samples) and \textbf{best-of-$N$} (maximum score among samples) aggregations. 

\paragraph{Evaluation Target.}
During training, we provide the model with the first $p\%$ of the ground truth $y^\star$ as an assistant prefix to seed generation (see \Cref{tab:grpo-dataset-hparams} for dataset-specific values). At evaluation time, we run two evaluation protocols:
\begin{enumerate}
    \item \textbf{Suffix-only evaluation}: We truncate the first $p\%$ of words from $y^\star$ before computing similarity, evaluating only the model's ability to reconstruct the \emph{novel} suffix portion that was not provided as a prefix.
    \item \textbf{Full evaluation}: We evaluate against the complete $y^\star$ (prefix + suffix), measuring overall reconstruction quality.
\end{enumerate}
The suffix-only evaluation provides a stricter measure of memorization, as it excludes the ``free'' tokens the model was given. Specifically, we report reconstruction quality under the suffix-only evaluation .

\paragraph{Metric Definitions.}
\Cref{tab:eval-metrics} summarizes the evaluation metrics. We adopt consistent notation with the training rewards (\Cref{tab:lexical-rewards}), where $\hat{y}$ denotes the model's completion and $y^\star$ denotes the ground truth. Let $T(\cdot)$ denote the set of unique tokens in a string, $\mathrm{LCS}(\cdot,\cdot)$ denote the longest common subsequence length, and $\mathcal{N}_{[L_{\min},L_{\max}]}(\cdot)$ denote the set of contiguous $n$-grams with $n \in [L_{\min}, L_{\max}]$.

\begin{table}[t]
\centering
\caption{\textbf{Evaluation metrics} for measuring reconstruction quality and producing membership score. Each metric is computed for both average-of-$N$ and best-of-$N$ aggregations.}
\label{tab:eval-metrics}
\begin{tabular}{@{}lll@{}}
\toprule
Metric & Definition & Normalization \\ 
\midrule
\multicolumn{3}{@{}l}{\textit{Token-based Metrics}} \\[0.3ex]
Jaccard similarity &
$\displaystyle s_{\text{jacc}}(\hat{y},y^\star) = \frac{\lvert T(\hat{y}) \cap T(y^\star)\rvert}{\lvert T(\hat{y}) \cup T(y^\star)\rvert}$ & 
Union (symmetric) \\[1.2ex]
Token overlap (ref) &
$\displaystyle s_{\text{tok}}(\hat{y},y^\star) = \frac{\lvert T(\hat{y}) \cap T(y^\star)\rvert}{\lvert T(y^\star)\rvert}$ & 
Reference \\[1.2ex]
Token overlap (cand) &
$\displaystyle s_{\text{tok}}^{\text{cand}}(\hat{y},y^\star) = \frac{\lvert T(\hat{y}) \cap T(y^\star)\rvert}{\lvert T(\hat{y})\rvert}$ & 
Candidate \\[0.8ex]
\midrule
\multicolumn{3}{@{}l}{\textit{Sequence-based Metrics}} \\[0.3ex]
LCS length &
$\displaystyle \mathrm{LCS}(\hat{y},y^\star)$ & 
Raw (unnormalized) \\[0.8ex]
LCS ratio (ref) &
$\displaystyle s_{\text{lcs}}(\hat{y},y^\star) = \frac{\mathrm{LCS}(\hat{y},y^\star)}{\lvert y^\star\rvert}$ & 
Reference \\[1.2ex]
LCS ratio (cand) &
$\displaystyle s_{\text{lcs}}^{\text{cand}}(\hat{y},y^\star) = \frac{\mathrm{LCS}(\hat{y},y^\star)}{\lvert \hat{y}\rvert}$ & 
Candidate \\[0.8ex]
\midrule
\multicolumn{3}{@{}l}{\textit{N-gram Metrics}} \\[0.3ex]
N-gram coverage (cand) &
$\displaystyle s_{\text{ng}}^{\text{cand}}(\hat{y},y^\star) = \frac{\big\lvert \mathcal{N}(\hat{y})\cap \mathcal{N}(y^\star)\big\rvert}{\big\lvert \mathcal{N}(\hat{y})\big\rvert}$ & 
Candidate \\[1.2ex]
N-gram coverage (ref) &
$\displaystyle s_{\text{ng}}(\hat{y},y^\star) = \frac{\big\lvert \mathcal{N}(\hat{y})\cap \mathcal{N}(y^\star)\big\rvert}{\big\lvert \mathcal{N}(y^\star)\big\rvert}$ & 
Reference \\[0.8ex]
\midrule
\multicolumn{3}{@{}l}{\textit{Semantic Metrics}} \\[0.3ex]
Embedding similarity &
$\displaystyle s_{\text{emb}}(\hat{y},y^\star) = \frac{\phi(\hat{y}) \cdot \phi(y^\star)}{\|\phi(\hat{y})\| \|\phi(y^\star)\|}$ & 
Cosine similarity \\[0.8ex]
\bottomrule
\end{tabular}%
\end{table}

\paragraph{Metric Details.}
\begin{itemize}
    \item \emph{Token-based metrics} measure vocabulary overlap between the completion and ground truth. Jaccard similarity normalizes by the union of token sets, penalizing both missing tokens and extraneous tokens equally. Token overlap metrics normalize by either the reference (measuring recall of ground truth tokens) or candidate (measuring precision of generated tokens). Reference-normalized metrics are more suitable for membership inference as they measure what fraction of the target was successfully reconstructed.
    
    \item \emph{Sequence-based metrics} capture ordering information via the longest common subsequence. The raw LCS length provides an absolute measure of matched content, while normalized variants express this as a fraction of reference or candidate length.
    
    \item \emph{N-gram metrics} measure the coverage of contiguous token sequences. Following prior work~\citep{hallinan2025surprising}, we compute $n$-grams for $n \in [3, \mathrm{LCS}(\hat{y}, y^\star)]$, treating $n$-grams as sets to discourage reward hacking via repetition.
    
    \item \emph{Embedding similarity} provides a semantic measure using Qwen3-Embedding-8B~\citep{zhang2025qwen3}. The cosine similarity between embedding vectors $\phi(\cdot)$ captures meaning beyond lexical overlap.
\end{itemize}

\paragraph{Reference vs.\ Candidate Normalization.}
Reference-normalized and candidate-normalized metrics capture complementary information. \emph{Reference-normalized} metrics (normalizing by $|y^\star|$ or $|T(y^\star)|$) measure the fraction of the ground truth that was successfully reconstructed, reflecting recall. \emph{Candidate-normalized} metrics (normalizing by $|\hat{y}|$ or $|T(\hat{y})|$) measure what fraction of the generated output matches the target, reflecting precision. We report both variants. In our RL training (\Cref{sec:app-method-details}), we use reference-normalized metrics as rewards, as they cannot be trivially maximized by generating shorter outputs.

\subsection{N-Sampling Baseline Evaluation}

For the N-sampling baseline, we generate $N$ independent rollouts for each input and compute the metrics defined above. To ensure a fair comparison with our RL-trained models, we match the sampling configuration used during RL training: temperature $\tau = 1.0$, top-$p = 0.95$, and top-$k = 50$. The number of rollouts $N$ also matches the rollout budget used in RL training (see \Cref{tab:grpo-dataset-hparams}).

\subsection{ADRA Evaluation}

After RL training, we evaluate the trained policy by generating $N$ rollouts per input, matching the rollout budget used in $N$-sampling and RL training for fair comparison. We reduce the sampling temperature to $\tau = 0.7$ while keeping top-$p = 0.95$ and top-$k = 50$. The lower temperature reflects that the RL-trained policy has already learned to generate reconstructions, so we bias toward higher-likelihood outputs rather than encouraging exploration. This also follows typical practice in RL evaluation, where exploitation is preferred over exploration at test time.




\section{Additional Results: Pre-training Reconstruction}
\label{sec:app-pre-train-reconstruction}
\begin{table*}[t]
\centering
\caption{
\textbf{Pre-training member data reconstruction} for paraphrased setting. \textbf{Bold} denotes the best average performance for each dataset. ADRA+ and ADRA consistently outperform the N-Sampling across all metrics. See \Cref{tab:mia-reconstruction-pretraining} for original verbatim setting results.
}
\label{tab:mia-reconstruction-pretraining-paraphrase}
\begingroup
\renewcommand{\sect}[1]{\rowcolor{gray!18}\multicolumn{8}{l}{\textbf{#1}}\\}
\resizebox{0.9\linewidth}{!}{%
\begin{tabular}{@{}lllccccc@{}}
\toprule
\textbf{Model} & \textbf{Type} & \textbf{Method} &
\makecell{\textbf{Budget} \\ (\textbf{$N$})} &
\makecell{\textbf{Lexical Jaccard}\\(Best / Avg)} &
\makecell{\textbf{Lexical LCS}\\ (Best / Avg)} &
\makecell{\textbf{Lexical Coverage}\\ (Best / Avg)} &
\makecell{\textbf{Embedding Cosine}\\ (Best / Avg)} \\
\midrule
\sect{BookMIA}
\addlinespace[0.6ex]
\multirow{4}{*}{Llama2-7B} & \multirow{4}{*}{Para.}
  & ADRA+ & 32 & \bestavg{15.8}{\textbf{13.5}} & \bestavg{51}{\textbf{44}} & \bestavg{3.0}{\textbf{1.0}} & \bestavg{88.2}{84.2} \\
\cmidrule(lr){3-8}
  &  & ADRA & 32 & \bestavg{14.7}{13.0} & \bestavg{47}{42} & \bestavg{2.2}{0.9} & \bestavg{87.3}{\textbf{84.3}} \\
\cmidrule(lr){3-8}
  &  & N-Sampling & 32 & \bestavg{15.5}{12.3} & \bestavg{48}{39} & \bestavg{2.5}{0.5} & \bestavg{88.2}{81.5} \\
\sect{WikiMIA24-Hard}
\addlinespace[0.6ex]
\multirow{4}{*}{Qwen2-7B} & \multirow{4}{*}{Para.}
  & ADRA+ & 32 & \bestavg{16.4}{13.1} & \bestavg{33}{25} & \bestavg{8.1}{4.0} & \bestavg{86.5}{82.0} \\
\cmidrule(lr){3-8}
  & & ADRA & 32 & \bestavg{17.3}{\textbf{13.4}} & \bestavg{40}{27} & \bestavg{9.5}{4.2} & \bestavg{88.1}{\textbf{84.1}} \\
\cmidrule(lr){3-8}
  &  & N-Sampling & 32 & \bestavg{14.3}{9.9} & \bestavg{47}{\textbf{37}} & \bestavg{12.8}{\textbf{5.3}} & \bestavg{87.2}{80.0} \\
\sect{Olmo3 Training Mix}
\addlinespace[0.6ex]
\multirow{4}{*}{Olmo3-7B-Instruct} & \multirow{4}{*}{Para.}
  & ADRA+ & 32 & \bestavg{14.3}{11.8} & \bestavg{56}{\textbf{48}} & \bestavg{6.3}{2.9} & \bestavg{86.3}{83.1} \\
\cmidrule(lr){3-8}
  & & ADRA & 32 & \bestavg{14.6}{\textbf{12.0}} & \bestavg{56}{47} & \bestavg{6.4}{\textbf{3.0}} & \bestavg{86.5}{\textbf{83.2}} \\
\cmidrule(lr){3-8}
  &  & N-Sampling & 32 & \bestavg{13.8}{10.2} & \bestavg{40}{25} & \bestavg{6.2}{1.6} & \bestavg{86.8}{82.5} \\
\bottomrule
\end{tabular}
}
\endgroup
\end{table*}


\Cref{tab:mia-reconstruction-pretraining-paraphrase} presents reconstruction quality for paraphrased members across pre-training datasets. Results closely mirror the verbatim setting (\Cref{tab:mia-reconstruction-pretraining}). \Method~and \MethodPlus~consistently outperform N-Sampling on average across lexical Jaccard and embedding cosine metrics. On BookMIA, \MethodPlus~achieves the strongest average performance (13.5 Jaccard, 44 LCS, 1.0 coverage), while \Method~attains the best embedding cosine (84.3). On WikiMIA$_{\text{2024}}$ Hard, \Method~reaches the best average Jaccard (13.4) and embedding cosine (84.1), though N-Sampling achieves higher LCS (37) and coverage (5.3)—consistent with the metric trade-off observed in the verbatim setting where RL optimization favors reward-aligned metrics. On Olmo3 Training Mix, \Method~attains the best average Jaccard (12.0), coverage (3.0), and embedding cosine (83.2), while \MethodPlus~achieves the highest LCS (48). Overall, results suggest that RL improves training data extraction despite semantic perturbations.

\section{Additional Results: Post-training Reconstruction}
\label{sec:app-post-train-reconstruction}
\begin{table*}[t]
\centering
\caption{
\textbf{Post-training member data reconstruction} for paraphrased setting. ADRA+ and ADRA consistently outperform the N-Sampling across both lexical and semantic metrics. See \Cref{tab:mia-post-training-data-reconstruction} for original verbatim setting results.
}
\label{tab:mia-post-training-data-reconstruction-paraphrase}
\begingroup
\renewcommand{\sect}[1]{\rowcolor{gray!18}\multicolumn{8}{l}{\textbf{#1}}\\}
\resizebox{0.9\linewidth}{!}{%
\begin{tabular}{@{}lllccccc@{}}
\toprule
\textbf{Model} & \textbf{Type} & \textbf{Method} &
\makecell{\textbf{Budget} \\ (\textbf{$N$})} &
\makecell{\textbf{Lexical Jaccard}\\(Best / Avg)} &
\makecell{\textbf{Lexical LCS}\\ (Best / Avg)} &
\makecell{\textbf{Lexical Coverage}\\ (Best / Avg)} &
\makecell{\textbf{Embedding Cosine}\\ (Best / Avg)} \\
\midrule
\sect{Olympia Math}
\addlinespace[0.6ex]
\multirow{4}{*}{Tulu2-7B} & \multirow{4}{*}{Para.}
  & ADRA+ & 32 & \bestavg{23.8}{21.0} & \bestavg{67}{60} & \bestavg{16.0}{\textbf{11.4}} & \bestavg{91.7}{89.3} \\
  \cmidrule(lr){3-8}
  &  & ADRA & 32 & \bestavg{24.1}{\textbf{21.4}} & \bestavg{68}{\textbf{61}} & \bestavg{12.7}{8.5} & \bestavg{92.1}{\textbf{89.6}} \\
\cmidrule(lr){3-8}
  &  & N-Sampling & 32 & \bestavg{22.8}{15.5} & \bestavg{60}{34} & \bestavg{11.9}{5.1} & \bestavg{92.4}{87.3} \\
\sect{AIME}
\addlinespace[0.6ex]
\multirow{4}{*}{Tulu2-7B} & \multirow{4}{*}{Para.}
  &  ADRA+ & 32 & \bestavg{22.7}{\textbf{18.8}} & \bestavg{54}{40} & \bestavg{17.5}{\textbf{12.7}} & \bestavg{92.2}{88.8} \\
  \cmidrule(lr){3-8}
  & & ADRA & 32 & \bestavg{21.1}{17.9} & \bestavg{56}{\textbf{42}} & \bestavg{15.4}{11.7} & \bestavg{91.8}{\textbf{89.0}} \\
\cmidrule(lr){3-8}
  &  & N-Sampling & 32 & \bestavg{18.4}{12.1} & \bestavg{48}{27} & \bestavg{13.6}{4.8} & \bestavg{89.2}{83.3} \\
\bottomrule
\end{tabular}
}
\endgroup
\end{table*}

\Cref{tab:mia-post-training-data-reconstruction-paraphrase} presents reconstruction quality for paraphrased members in post-training settings. Results closely mirror the verbatim setting (\Cref{tab:mia-post-training-data-reconstruction}) with better extraction compared to pre-training.  \Method~and \MethodPlus~consistently outperform N-Sampling across all metrics on both controlled contamination datasets. On Olympia Math, \Method~achieves the best average Jaccard (21.4), LCS (61), and embedding cosine (89.6), while \MethodPlus~attains the highest coverage (11.4). On AIME, \Method~excels in average Jaccard (18.8) and Coverage (12.7) while \Method excels in average LCS (42) and embedding cosine (89.0). The improvements over N-Sampling remain substantial: on AIME, \Method~improves average Jaccard by 6.7, LCS by 15, coverage by 7.9, and embedding cosine by 5.7. Like in pre-training, results confirms that RL improves training data extraction despite semantic perturbations.

\section{Reconstructions Examples}

We show qualitative reconstruction examples across all three training stages: pre-training (\Cref{fig:qual_example_arxiv}, Olmo3 Mix arXiv), post-training (\Cref{fig:qual_example_1,fig:qual_example_2}, AIME), and distillation (\Cref{fig:qual_example_s1_r1}, S1.1 Deepseek-R1).


\definecolor{SkinColor}{RGB}{253, 247, 244}
\definecolor{SkyBlue}{RGB}{103, 143, 195}
\definecolor{DarkSkyBlue}{RGB}{70, 100, 140}
\definecolor{LightGreen}{RGB}{200, 230, 201}
\definecolor{LightBlue}{RGB}{187, 222, 251}
\definecolor{HighlightYellow}{RGB}{255, 249, 196}
\begin{figure}[t]
\centering
\begin{tcolorbox}[
    colframe=DarkSkyBlue!80!black,
    colback=LightGreen,
    coltitle=white,
    colbacktitle=DarkSkyBlue!80!black,
    title=\textbf{Ground Truth},
    fonttitle=\bfseries,
    boxsep=3pt,
    left=6pt,
    right=6pt,
    top=3pt,
    bottom=3pt,
    arc=2pt,
    width=\textwidth
]
\small
The novelty of our approach is in \colorbox{HighlightYellow}{probabilistically modelling the HT as a continuous function of frequency,} rather than using multiple independent response models (psychometric functions) for a discrete set of standard frequencies. This is by endowing the \colorbox{HighlightYellow}{threshold curve} with a \colorbox{HighlightYellow}{Gaussian process (GP) prior} \textbackslash citep\{rasmussen\_gaussian\_2006\}, which is why we refer to our method as `GP-PTA'. If the responses are binary (audible or non-audible), the full model can be interpreted as a two-dimensional binary classifier that is specified by a GP in the frequency dimension and by a psychometric (sigmoid) function in the intensity dimension. It provides \colorbox{HighlightYellow}{uncertainty bands on the resulting threshold estimate}, which enables fundamental and objective \colorbox{HighlightYellow}{stopping criteria.} Moreover, it allows one to find the \colorbox{HighlightYellow}{optimal next stimulus} based on the data processed so far, and stimuli at any frequency can be used. These properties \colorbox{HighlightYellow}{reduce the number of} trials \colorbox{HighlightYellow}{required} to achieve the desired accuracy. Minimizing the amount or required trials is important to reduce the cognitive burden on the (often elderly) patients. [\ldots] A probabilistic response model is specified to capture uncertainty in the patient's responses [\ldots] A generative probabilistic model is derived by combining the response model with a GP. The \colorbox{HighlightYellow}{GP provides} a natural way to take \colorbox{HighlightYellow}{prior knowledge about the threshold curve into account,} and relaxes the piecewise linear assumption for the \colorbox{HighlightYellow}{threshold curve} to a more natural \colorbox{HighlightYellow}{smoothness assumption.}
\end{tcolorbox}
\vspace{0.3cm}
\begin{tcolorbox}[
    colframe=SkyBlue,
    colback=LightBlue,
    coltitle=white,
    colbacktitle=SkyBlue,
    title=\textbf{Reconstruction},
    fonttitle=\bfseries,
    boxsep=3pt,
    left=6pt,
    right=6pt,
    top=3pt,
    bottom=3pt,
    arc=2pt,
    width=\textwidth
]
\small
However, these approaches typically assume that the \colorbox{HighlightYellow}{hearing thresholds at different frequencies} are independently distributed [\ldots] Recent work using \colorbox{HighlightYellow}{Gaussian process (GP)} models \textbackslash citep\{zhang2017gp, garrido2019gp, rahman2020gp\} has begun to address this by \colorbox{HighlightYellow}{modeling the HT curve as a Gaussian process}, which naturally captures correlations between adjacent frequency points. These models are capable of providing not only point estimates for the threshold at each frequency but also \colorbox{HighlightYellow}{uncertainty estimates,} which can be used to determine an objective \colorbox{HighlightYellow}{stopping criterion} [\ldots] \colorbox{HighlightYellow}{GP} models can be regularized with \colorbox{HighlightYellow}{prior information about} typical HT curves [\ldots] In this work, we propose a simplified yet effective framework for estimating the pure tone hearing \colorbox{HighlightYellow}{threshold curve} using a structured Gaussian process. Our approach models the entire threshold curve as a \colorbox{HighlightYellow}{function of frequency} using a \colorbox{HighlightYellow}{covariance function} that explicitly encodes \colorbox{HighlightYellow}{smoothness} and periodicity constraints [\ldots] This structure allows the model to extrapolate between measured points and to exploit the inherent correlation between thresholds at neighboring frequencies, thus \colorbox{HighlightYellow}{reducing the number of required measurements.} We also incorporate a prior based on standardized audiograms [\ldots] A method to use the output of the model for \colorbox{HighlightYellow}{adaptive data collection,} including a \colorbox{HighlightYellow}{stopping criterion} based on the uncertainty of the estimated thresholds.
\end{tcolorbox}
\caption{Qualitative example from Olmo3 Mix Arxiv: Reconstruction of a GP-based audiometry paper. Highlighted phrases show semantic and lexical overlap in core technical concepts---\textit{Gaussian process (GP)}, \textit{threshold curve}, \textit{function of frequency}, \textit{uncertainty estimates/bands}, \textit{stopping criterion/criteria}, \textit{smoothness}, and \textit{prior knowledge/information}---suggesting that the reconstruction recovers the paper's key methodological vocabulary and conceptual structure.}
\label{fig:qual_example_arxiv}
\end{figure}
\definecolor{SkinColor}{RGB}{253, 247, 244}
\definecolor{SkyBlue}{RGB}{103, 143, 195}
\definecolor{DarkSkyBlue}{RGB}{70, 100, 140}
\definecolor{LightGreen}{RGB}{200, 230, 201}
\definecolor{LightBlue}{RGB}{187, 222, 251}
\definecolor{HighlightYellow}{RGB}{255, 249, 196}

\begin{figure}[t]
\centering

\begin{tcolorbox}[
    colframe=DarkSkyBlue!80!black,
    colback=LightGreen,
    coltitle=white,
    colbacktitle=DarkSkyBlue!80!black,
    title=\textbf{Ground Truth},
    fonttitle=\bfseries,
    boxsep=3pt,
    left=6pt,
    right=6pt,
    top=3pt,
    bottom=3pt,
    arc=2pt,
    width=\textwidth
]
\small
\colorbox{HighlightYellow}{$d \cdot e \cdot f =$} $(a \cdot b \cdot c) \cdot (a \cdot b \cdot c - 1) \geq 6! = 720,$ so $a \cdot b \cdot c \geq 28.$ It follows that $a \cdot b \cdot c$ and $d \cdot e \cdot f$ are consecutive composites with prime factors no other than $2,3,5,$ and $7.$ The smallest values for $a \cdot b \cdot c$ and $d \cdot e \cdot f$ are $36$ and $35,$ respectively. So, we have $\{a,b,c\} = \{2,3,6\}, \{d,e,f\} = \{1,5,7\},$ and $\{g,h,i\} = \{4,8,9\},$ from which \colorbox{HighlightYellow}{$\frac{a \cdot b \cdot c - d \cdot e \cdot f}{g \cdot h \cdot i} =$} $\frac{1}{288}.$ If we do not \colorbox{HighlightYellow}{minimize the numerator}, then $a \cdot b \cdot c - d \cdot e \cdot f > 1.$ Note that \colorbox{HighlightYellow}{$\frac{a \cdot b \cdot c - d \cdot e \cdot f}{g \cdot h \cdot i}$} $\geq \frac{2}{7\cdot8\cdot9} > \frac{1}{288}.$Together, we conclude that \colorbox{HighlightYellow}{the minimum possible positive value of $\frac{a \cdot b \cdot c - d \cdot e \cdot f}{g \cdot h \cdot i}$ is} $\frac{1}{288}.$ \colorbox{HighlightYellow}{Therefore, the answer is} $1+288=\boxed{289}.$
\end{tcolorbox}

\vspace{0.3cm}

\begin{tcolorbox}[
    colframe=SkyBlue,
    colback=LightBlue,
    coltitle=white,
    colbacktitle=SkyBlue,
    title=\textbf{Reconstruction},
    fonttitle=\bfseries,
    boxsep=3pt,
    left=6pt,
    right=6pt,
    top=3pt,
    bottom=3pt,
    arc=2pt,
    width=\textwidth
]
\small
\colorbox{HighlightYellow}{$d \cdot e \cdot f = $}$1.$ \colorbox{HighlightYellow}{Therefore, the minimum possible positive value of $\frac{a \cdot b \cdot c - d \cdot e \cdot f}{g \cdot h \cdot i}$ is} $\frac{1}{7\cdot8\cdot9}.$

Similarly, to minimize a positive fraction, we minimize its numerator and maximize its denominator. It is clear that \colorbox{HighlightYellow}{$\frac{a \cdot b \cdot c - d \cdot e \cdot f}{g \cdot h \cdot i} \geq$} $\frac{1}{7\cdot8\cdot9}.$ If we \colorbox{HighlightYellow}{minimize the numerator}, then $a \cdot b \cdot c - d \cdot e \cdot f = 1.$ Note that $a \cdot b \cdot c \cdot d \cdot e \cdot f = 1.$ \colorbox{HighlightYellow}{Therefore, the minimum possible positive value of $\frac{a \cdot b \cdot c - d \cdot e \cdot f}{g \cdot h \cdot i}$ is} $\frac{1}{7\cdot8\cdot9}.$

Therefore, the sum of the relatively prime positive integers is $1+1=2.$

Therefore, we need to find \colorbox{HighlightYellow}{the minimum possible positive value of $\frac{a \cdot b \cdot c - d \cdot e \cdot f}{g \cdot h \cdot i}.$} In this case, \colorbox{HighlightYellow}{the minimum possible positive value is} $\frac{1}{7\cdot8\cdot9}.$

\colorbox{HighlightYellow}{Therefore, the answer is}$\frac{1}{7\cdot8\cdot9}.$

\colorbox{HighlightYellow}{Therefore, the answer is} $2.$
\end{tcolorbox}

\caption{Qualitative Example 1 from AIME. The reconstruction captures core mathematical reasoning patterns and expressions from the ground truth, despite yielding wrong final answer.}
\label{fig:qual_example_1}
\end{figure}
\begin{figure}[t]
\centering

\begin{tcolorbox}[
    colframe=DarkSkyBlue!80!black,
    colback=LightGreen,
    coltitle=white,
    colbacktitle=DarkSkyBlue!80!black,
    title=\textbf{Ground Truth},
    fonttitle=\bfseries,
    boxsep=3pt,
    left=6pt,
    right=6pt,
    top=3pt,
    bottom=3pt,
    arc=2pt,
    width=\textwidth
]
\small
before the bus arrives \colorbox{HighlightYellow}{are adults. Similarly,} we know that $x + 50 \equiv 0 \pmod{25}$, \colorbox{HighlightYellow}{as $\frac{11}{25}$ of} the \colorbox{HighlightYellow}{people at the party are adults after the bus arrives.} $x + 50 \equiv 0 \pmod{25}$ can be reduced to $x \equiv 0 \pmod{25}$, and since we are looking for the minimum amount of people, $x$ is $300$. That means there are $350$ people \colorbox{HighlightYellow}{at the party after the bus arrives,} and thus there are $350 \cdot \frac{11}{25} = \boxed{154}$ adults at the party.
\end{tcolorbox}

\vspace{0.3cm}

\begin{tcolorbox}[
    colframe=SkyBlue,
    colback=LightBlue,
    coltitle=white,
    colbacktitle=SkyBlue,
    title=\textbf{Reconstruction},
    fonttitle=\bfseries,
    boxsep=3pt,
    left=6pt,
    right=6pt,
    top=3pt,
    bottom=3pt,
    arc=2pt,
    width=\textwidth
]
\small
\colorbox{HighlightYellow}{are adults.}

\colorbox{HighlightYellow}{Similarly,} let $x$ be the number of people at the party before the bus cut. We know that $x\equiv 0\pmod{12}$, \colorbox{HighlightYellow}{as $\frac{5}{12}$ of people at the party are adults.}

Therefore, the number of adults at the party before the bus arrives is $0$.

After the bus arrives, there are a total of $50+25=75$ \colorbox{HighlightYellow}{people at the party}. Let $x$ be the number of \colorbox{HighlightYellow}{people at the party after the bus arrives.} We know that $x\equiv 0\pmod{12}$, \colorbox{HighlightYellow}{as $\frac{11}{25}$ of people at the party are adults.}

Therefore, the minimum number of \colorbox{HighlightYellow}{adults at the party after the bus arrives} is $0$.

Therefore, the answer is $0$.

We are looking for the minimum number of \colorbox{HighlightYellow}{adults at the party after the bus arrives.} In this case, there are no \colorbox{HighlightYellow}{adults at the party after the bus arrives.}

Therefore, the answer is $0$.
\end{tcolorbox}

\caption{Qualitative Example 2 from AIME. The reconstruction captures key mathematical patterns and modular arithmetic reasoning despite different final answer.}
\label{fig:qual_example_2}
\end{figure}
\definecolor{SkinColor}{RGB}{253, 247, 244}
\definecolor{SkyBlue}{RGB}{103, 143, 195}
\definecolor{DarkSkyBlue}{RGB}{70, 100, 140}
\definecolor{LightGreen}{RGB}{200, 230, 201}
\definecolor{LightBlue}{RGB}{187, 222, 251}
\definecolor{HighlightYellow}{RGB}{255, 249, 196}

\begin{figure}[t]
\centering
\begin{tcolorbox}[
    colframe=DarkSkyBlue!80!black,
    colback=LightGreen,
    coltitle=white,
    colbacktitle=DarkSkyBlue!80!black,
    title=\textbf{Ground Truth},
    fonttitle=\bfseries,
    boxsep=3pt,
    left=6pt,
    right=6pt,
    top=3pt,
    bottom=3pt,
    arc=2pt,
    width=\textwidth
]
\small
\colorbox{HighlightYellow}{Given the matrix $\mathbf{M} = \left(\begin{smallmatrix} -\omega^2 & -\omega \\ 1 & 0 \end{smallmatrix}\right)$, where $\omega$ is a nonreal root of $x^3 = 1$,} 

\colorbox{HighlightYellow}{we need to find the sum of the entries of $\mathbf{M} + \mathbf{M}^2 + \mathbf{M}^3 + \dots + \mathbf{M}^{2009}$.}

\colorbox{HighlightYellow}{First, we note that} $\omega$ is a primitive cube root of unity, satisfying \colorbox{HighlightYellow}{$\omega^3 = 1$ and $1 + \omega + \omega^2 = 0$.} We \colorbox{HighlightYellow}{compute the} \colorbox{HighlightYellow}{powers of $\mathbf{M}$:}

1. $\mathbf{M} = \left(\begin{smallmatrix} -\omega^2 & -\omega \\ 1 & 0 \end{smallmatrix}\right)$\quad
2. $\mathbf{M}^2 = \left(\begin{smallmatrix} 0 & 1 \\ -\omega^2 & -\omega \end{smallmatrix}\right)$\quad
3. \colorbox{HighlightYellow}{$\mathbf{M}^3 = \left(\begin{smallmatrix} 1 & 0 \\ 0 & 1 \end{smallmatrix}\right)$} (the identity matrix)

\colorbox{HighlightYellow}{Since $\mathbf{M}^3 = \mathbf{I}$, the powers of $\mathbf{M}$ cycle every 3 terms.} We decompose the sum [\ldots] into cycles of 3 terms. There are \colorbox{HighlightYellow}{$2009$ terms} in total, which is \colorbox{HighlightYellow}{$669$ complete cycles} (each containing 3 terms) and \colorbox{HighlightYellow}{2 remaining terms.}

The sum for each cycle is $\mathbf{M} + \mathbf{M}^2 + \mathbf{I}$. The remaining terms are \colorbox{HighlightYellow}{$\mathbf{M}^{2008}$} and \colorbox{HighlightYellow}{$\mathbf{M}^{2009}$}.

\colorbox{HighlightYellow}{The sum of the entries} of [\ldots] is computed as follows: - \colorbox{HighlightYellow}{Sum of entries of $\mathbf{M}$:} $-\omega^2 - \omega + 1 + 0 = 1 - \omega^2 - \omega$ -
\colorbox{HighlightYellow}{Sum of entries of $\mathbf{M}^2$:} $0 + 1 - \omega^2 - \omega = 1 - \omega^2 - \omega$ - \colorbox{HighlightYellow}{Sum of entries of $\mathbf{I}$: $1 + 0 + 0 + 1 = 2$}

Total sum for one cycle: $2(1 - \omega^2 - \omega) + 2 = 6$. [\ldots] \colorbox{HighlightYellow}{Therefore, the total sum is:} $669 \cdot 6 + 4 = 4014 + 4 = 4018$

\colorbox{HighlightYellow}{The sum of the entries of $\mathbf{M} + \mathbf{M}^2 + \mathbf{M}^3 + \dots + \mathbf{M}^{2009}$ is }$\boxed{4018}$.
\end{tcolorbox}
\vspace{0.3cm}
\begin{tcolorbox}[
    colframe=SkyBlue,
    colback=LightBlue,
    coltitle=white,
    colbacktitle=SkyBlue,
    title=\textbf{Reconstruction},
    fonttitle=\bfseries,
    boxsep=3pt,
    left=6pt,
    right=6pt,
    top=3pt,
    bottom=3pt,
    arc=2pt,
    width=\textwidth
]
\small
\colorbox{HighlightYellow}{Given the matrix $\mathbf{M} = \left(\begin{smallmatrix} -\omega^2 & -\omega \\ 1 & 0 \end{smallmatrix}\right)$, where $\omega$ is a nonreal root of $x^3 = 1$,}

\colorbox{HighlightYellow}{we need to find the sum of the entries of $\mathbf{M} + \mathbf{M}^2 + \mathbf{M}^3 + \dots + \mathbf{M}^{2009}$.}

\colorbox{HighlightYellow}{First, we note that} the nonreal roots of $x^3 = 1$ are $\omega$ and $\omega^2$, where $\omega$ and $\omega^2$ are the complex cube roots of unity satisfying \colorbox{HighlightYellow}{$\omega^3 = 1$ and $1 + \omega + \omega^2 = 0$.}

To find the \colorbox{HighlightYellow}{powers of $\mathbf{M}$,} we \colorbox{HighlightYellow}{compute the} first few powers of $\mathbf{M}$:
$\mathbf{M}^2 = \left(\begin{smallmatrix} \omega & 1 \\ -\omega^2 & -\omega \end{smallmatrix}\right)$\quad
\colorbox{HighlightYellow}{$\mathbf{M}^3 = \left(\begin{smallmatrix} 1 & 0 \\ 0 & 1 \end{smallmatrix}\right) = \mathbf{I}$}

\colorbox{HighlightYellow}{Since $\mathbf{M}^3 = \mathbf{I}$, the powers of $\mathbf{M}$ cycle every 3 terms.} Therefore, the sequence $\mathbf{M}, \mathbf{M}^2, \mathbf{M}^3, \mathbf{M}^4, \ldots$ repeats every 3 terms.

Next, we find \colorbox{HighlightYellow}{the sum of the entries} of each matrix in one cycle: - \colorbox{HighlightYellow}{The sum of the entries of $\mathbf{M}$} is $(-\omega^2) + (-\omega) + 1 + 0 = 1 - \omega - \omega^2 = 0$. - \colorbox{HighlightYellow}{The sum of the entries of $\mathbf{M}^2$} is $\omega + 1 - \omega^2 - \omega = 1$. - \colorbox{HighlightYellow}{The sum of the entries of $\mathbf{I}$ is $1 + 0 + 0 + 1 = 2$.}

Thus, the sum of the entries in one complete cycle (3 terms) is $0 + 1 + 2 = 3$.

[\ldots] we note that there are \colorbox{HighlightYellow}{$2009$ terms,} which can be divided into \colorbox{HighlightYellow}{$669$ complete cycles} of 3 terms and \colorbox{HighlightYellow}{2 remaining terms.} The sum of the entries for the complete cycles is $669 \times 3 = 2007$. The sum of the entries for the remaining 2 terms (\colorbox{HighlightYellow}{$\mathbf{M}^{2008}$} and \colorbox{HighlightYellow}{$\mathbf{M}^{2009}$}) is $1 + 2 = 3$.

\colorbox{HighlightYellow}{Therefore, the total sum is:} $2007 + 3 = 2010$

Thus, \colorbox{HighlightYellow}{the sum of the entries of $\mathbf{M} + \mathbf{M}^2 + \mathbf{M}^3 + \dots + \mathbf{M}^{2009}$ is} $\boxed{2010}$.
\end{tcolorbox}
\caption{Qualitative example from S1.1 Deepseek-R1 Distillation. Near-verbatim reconstruction of a matrix powers problem. Highlighted parts show overlap in algebraic identities, the cyclic structure insight ($\mathbf{M}^3=\mathbf{I}$), cycle decomposition logic, and entry-sum computation steps. Despite closely reproducing the full solution template, the reconstruction arrives at an incorrect final answer ($2010$ vs.\ $4018$).}
\label{fig:qual_example_s1_r1}
\end{figure}


\section{Ablation Details}
\label{sec:app-ablation-details}

Due to computational constraints, all ablations are conducted on AIME using a single seed.

\paragraph{Model-Based Reward Implementation.}
We implement model-based rewards by hosting dedicated embedding and vLLM inference servers, which the \textit{verl} trainer queries via FastAPI. Specifically, we use Qwen3-8B-Embedding as the embedding model and Qwen3-32B as the LLM-as-judge.

For embedding-based rewards, the model embeds both the candidate and ground-truth suffix and returns their cosine similarity as the reward signal. For LLM-as-judge rewards, the model is prompted to assess similarity between the candidate and ground-truth based on a structured prompt and returns a numerical score. The prompt used for LLM-as-judge can be found in \Cref{fig:llm_judge_prompt}.

\paragraph{Observations.}
In preliminary experiments, we find that model-based rewards are particularly prone to reward hacking, with different target models exhibiting distinct hacking behaviors. Our ablations suggest that model-based reward RL training can improve reconstruction performance, but its efficacy depends on both prompt design and model choice. We leave further exploration of model-based rewards for future work.

\begin{figure}[t]
\centering
\begin{tcolorbox}[
    colframe=SkyBlue,
    colback=SkinColor,
    coltitle=white,
    colbacktitle=DarkSkyBlue,
    title=\textbf{LLM-as-Judge Prompt Template},
    fonttitle=\bfseries,
    boxsep=5pt,
    left=8pt,
    right=8pt,
    top=5pt,
    bottom=5pt,
    arc=3pt,
    width=\textwidth
]
\small
\textbf{TASK DESCRIPTION}

Rate two solutions to the same math problem (one reference, one candidate) for similarity. The final answer in each solution is enclosed in ``\textbackslash boxed\{\}''. Return a real-valued score between 0 and 1 with exactly 3 decimals.

\vspace{0.5em}
\textbf{EVALUATION CRITERIA (consider all, equally)}

\begin{enumerate}
\item \textbf{Mathematical correctness} --- Is the candidate mathematically sound as in the reference?
\item \textbf{Solution approach} --- Does the candidate use methods/reasoning similar to the reference?
\item \textbf{Final answer} --- Does the candidate arrive at the same \textbackslash boxed\{\} answer as the reference?
\item \textbf{Overall clarity} --- Are the candidate's reasoning and steps as consistent and logically sound as the reference?
\item \textbf{Surface \& semantic resemblance} --- Do the two solutions look very similar in wording, symbols, step order, and intermediate expressions (i.e., high lexical/token overlap and near-paraphrase semantics)? Penalize added filler or unnecessary rephrasing.
\end{enumerate}

\vspace{0.5em}
\textbf{INPUTS}

\begin{itemize}
\item Reference solution: \texttt{\{REFERENCE\_SOLUTION\}}
\item Candidate solution: \texttt{\{CANDIDATE\_SOLUTION\}}
\end{itemize}

\vspace{0.5em}
\textbf{OUTPUT FORMAT (must follow exactly)}

Output ONLY one line:

\texttt{REWARD: <real value between 0 and 1 with 3 decimals>}
\end{tcolorbox}
\caption{Prompts used for LLM-as-Judge reward model evaluation.}
\label{fig:llm_judge_prompt}
\end{figure}

\section{Limitations \& Discussions}

There are limitations to our method and analysis; we detail them below.

\paragraph{Compute} ADRA is much more compute-intensive than current MIAs because it requires on-policy RL training. Performing MIA on large-scale data may be prohibitively expensive. Further, while we find that RL on average improves MIA and reconstruction over $N$-sampling, the efficacy of \textit{scaling RL training} varies across datasets. In some seeds and datasets, MIA performance peaks during early RL training steps, and continued training could degrade the performance.

\paragraph{Memorization \& Generalization} We hypothesize that latent traces of training data exist in the model weights that can be elicited through RL training. However, it is difficult to measure precisely how much our RL formulation elicits latent memorization versus generalization, which is also dataset-dependent. Precisely disentangling memorization and generalization has been an ongoing challenge~\citep{tanzer2022memorisation, morris2025languagemodelsmemorize}, and we hope future work can more carefully examine these effects at scale.

\paragraph{RL Algorithms.} We use vanilla GRPO~\citep{shao2024deepseekmath} throughout the paper. Recent work~\citep{liu2025understanding, yu2025dapo, qi2025defeating} has identified failure modes of GRPO, such as training collapse and instability, and proposed several remedies. These improvements could be readily incorporated into ADRA to further boost reconstruction and MIA performance. We leave exploration of better RL algorithms to future work.

Overall, we find that ADRA, the first active MIA, consistently improves over passive MIAs, revealing that model weights encode substantially more information about their training data than passive methods can surface — a finding with important implications for both membership inference and memorization research.


\end{document}